\pdfoutput=1

\documentclass[11pt]{article}

\usepackage[preprint]{acl}

\usepackage{times}
\usepackage{latexsym}

\usepackage[T1]{fontenc}
\usepackage[utf8]{inputenc}

\usepackage{microtype}

\usepackage{inconsolata}

\usepackage{graphicx}
\usepackage{hyperref}
\usepackage{url}
\usepackage{booktabs}
\usepackage{subcaption}
\usepackage{xspace}
\usepackage{colortbl}
\usepackage{xcolor}
\usepackage{float}
\usepackage[most]{tcolorbox}
\usepackage{longtable}

\newcommand{\data}{\textsc{ManyICLBench}\xspace}

\title{On Many-Shot In-Context Learning for Long-Context Evaluation}

\author{Kaijian Zou, Muhammad Khalifa, \and Lu Wang\\
  Computer Science and Engineering, University of Michigan, Ann Arbor, MI\\
  \texttt{\{zkjzou,khalifam,wangluxy\}@umich.edu}}

\begin{document}
\maketitle
\begin{abstract}
Many-shot in-context learning (ICL) has emerged as a unique setup to both utilize and test the ability of large language models to handle long context.
This paper delves into long-context language model (LCLM) evaluation through many-shot ICL. We first ask: 
what types of ICL tasks benefit from additional demonstrations, and how effective are they in evaluating LCLMs?
We find that classification and summarization tasks show performance improvements with additional demonstrations, while translation and reasoning tasks do not exhibit clear trends. 
Next, we investigate the extent to which different tasks necessitate retrieval versus global context understanding.
We develop metrics to categorize ICL tasks into two groups: (i) similar-sample learning (\textbf{SSL}): tasks where retrieval of the most similar examples is sufficient for good performance, and (ii) all-sample learning (\textbf{ASL}): tasks that necessitate a deeper comprehension of all examples in the prompt. 
Lastly, we introduce a new many-shot ICL benchmark built on existing ICL tasks, \textbf{\data}, to characterize model's ability on both fronts and benchmark 12 LCLMs using \data. We find that while state-of-the-art models demonstrate good performance up to 64k tokens in SSL tasks, many models experience significant performance drops at only 16k tokens in ASL tasks.\footnote{Data and code are available at \url{https://github.com/launchnlp/ManyICLBench}.}
\end{abstract}

\section{Introduction}

Long-context language models (LCLMs) have revolutionized the way users interact with language models by extending the context size from 2K to 128K or even 1M tokens \citep{geminiteam2024geminifamilyhighlycapable, glm2024chatglmfamilylargelanguage, dubey2024llama3herdmodels}. This unlocks challenging applications, such as long- and multi-document summarization, multi-turn dialogue, and code repository comprehension. 
Despite the recent progress in building LCLMs, existing benchmarks primarily evaluate these models' retrieval capabilities \citep{liu2023lostmiddlelanguagemodels, hsieh2024rulerwhatsrealcontext}. From synthetic tasks such as Needle-in-A-Haystack (NIAH) \citep{needle-in-a-haystack} and RULER benchmark~\citep{hsieh2024rulerwhatsrealcontext} to real-world challenges like long-novel QA \citep{karpinska2024thousandpairsnovelchallenge}, the majority of benchmarks assess how well LCLMs retrieve specific pieces of information from extensive contexts. As a result, \textbf{evaluating models' global understanding of the full context remains lacking}. 

To fill the gap, \citet{li2024longcontextllmsstrugglelong} introduce LongICLBench, which uses many-shot ICL classification tasks to evaluate models' long-context performance, arguing that these tasks require the comprehension of the entire input. 
A few other works have also explored many-shot ICL for long-context models \citep{agarwal2024manyshotincontextlearning, bertsch2024incontextlearninglongcontextmodels}. 
However, they have mainly relied on classification tasks ~\citep{li2024longcontextllmsstrugglelong,bertsch2024incontextlearninglongcontextmodels}, which are insufficient to distinguish which skills LCLMs require to perform well in many-shot ICL classification tasks. \citet{agarwal2024manyshotincontextlearning} study non-classification ICL tasks but only limit its study to Gemini 1.5 Pro.
In this work, we want to conduct a comprehensive study on many-shot ICL across a wide range of models, with a goal of identifying tasks that \textbf{benefit from additional demonstrations} and explore their utility in evaluating long-context models. Moreover, we seek to determine the extent to which these tasks rely on \textbf{retrieving most similar examples versus learning from all samples}.
\paragraph{RQ1: Which tasks benefit from many-shot ICL?} First, we investigate ICL tasks that are used in prior work, including classification, summarization, and reasoning, under many-shot settings with context lengths from 1k to 128k~\citep{agarwal2024manyshotincontextlearning}. We find that classification and summarization tasks show \textit{strong positive correlation between context lengths and model performance}. Our findings indicate that translation and reasoning tasks such as ARC \citep{allenai:arc} and FLORES-200 \citep{nllb2022} do not gain much performance with an increasing number of demonstrations. 
Science and symbolic reasoning tasks exhibit inconsistent trends between context lengths and model performance. 
This variance in performance is mainly attributed to the specific nature of tasks, where more demonstrations do not boost the models' task understanding.
Interestingly, math tasks benefit from additional demonstrations only when step-by-step solutions (or chain-of-thoughts) are derived and using strong LCLMs.

\textbf{RQ2: To what extent does each task require learning from a limited number of samples versus learning from more samples with broader context from LCLMs?} 
We use the ratio between the performance change of removing dissimilar examples and the change of removing similar examples. 
A high ratio means a more pronounced drop in performance upon removing similar examples, which indicates the task's heavy reliance on retrieving and learning its prediction with more similar examples.
Our analysis indicates that existing many-shot ICL \textit{classification} tasks \citep{li2024longcontextllmsstrugglelong} predominantly assess models' skills to learn from \textit{similar examples} rather than all examples, overshadowing the model's ability to understand all samples. This leads us to categorize tasks into similar-sample learning (SSL) and all-sample learning groups (ASL).

Following the categorization, we collect existing tasks and propose a new many-shot ICL benchmark, \textbf{\data}, designed to evaluate long-context models and advocate for including many-shot ICL tasks as effective evaluation candidates. Importantly, on \data, models are tested to either retrieve and learn from the most similar demonstrations or assimilate and learn from all demonstrations to enhance their understanding of the task \citep{lin2024dualoperatingmodesincontext, bertsch2024incontextlearninglongcontextmodels}. 
Therefore, \textit{\data evaluates both retrieval skills and global context understanding}, thus providing a holistic assessment of long-context models' capabilities.

Unlike NIAH, which focuses on retrieving a statement given a query, \data redefines retrieval to test a model’s ability to identify and leverage similar examples during in-context learning. Additionally, the benchmark retains a focus on global context understanding, as explored in prior works \citep{zhang2024inftybenchextendinglongcontext, karpinska2024thousandpairsnovelchallenge}, by requiring models to reason over and synthesize information from all demonstrations to generate accurate outputs.

In summary, we make contributions as below:
\begin{itemize}
    \vspace{-2mm}
    \item Investigate whether ICL tasks benefit from additional demonstrations and assess their suitability for evaluating LCLMs with a context length up to 128k tokens.
    \vspace{-2mm}
    \item Develop methods to characterize the primary skills evaluated by ICL tasks, where we focus on distinguishing between similar-sample learning and all-sample learning skills.
    \vspace{-2mm}
    \item Construct a many-shot ICL benchmark, \data, designed for evaluating LCLMs on both retrieval and global context understanding, while excluding irrelevant datasets previously used in evaluation.
    \vspace{-2mm}
    \item Benchmark 12 widely-used state-of-the-art LCLMs on \data to assess their performance comprehensively.
\end{itemize}
\section{Related Work}
\label{related_work}

\subsection{Long-Context Language Models and Evaluation}
As large language models (LLMs) grow in scale, there is an increasing demand for handling tasks that require extended contexts. Tasks such as long document summarization \citep{kryściński2022booksumcollectiondatasetslongform}, conversations with long-context memory \citep{xu2021goldfishmemorylongtermopendomain}, and repository-level code completion \citep{zhang-etal-2023-repocoder} have garnered significant interest.
Advances in efficient attention mechanisms, such as flash attention \citep{dao2022flashattentionfastmemoryefficientexact} and grouped query attention \citep{ainslie2023gqatraininggeneralizedmultiquery}, alongside the development of GPUs with larger memory capacities, have enabled LLMs to be trained on extended contexts. Techniques like position interpolation \citep{chen2023extendingcontextwindowlarge, peng2023yarnefficientcontextwindow} and context compression \citep{chevalier2023adaptinglanguagemodelscompress,mohtashami2023landmarkattentionrandomaccessinfinite,jiang2024longllmlinguaacceleratingenhancingllms} have further extended the context window size to up to 1 million tokens.

Despite these advancements, the NLP community still seeks a universal and effective method for evaluating long-context models. One prominent task is Needle-in-a-Haystack \citep{needle-in-a-haystack}, which requires models to retrieve the most relevant document from a large set of documents. Currently, most evaluation benchmarks focus on synthetic tasks that primarily assess the retrieval capabilities of long-context models \citep{hsieh2024rulerwhatsrealcontext, needle-in-a-haystack, lee2024longcontextlanguagemodelssubsume,lei2024s3evalsyntheticscalablesystematic}. Only a few benchmarks, such as \citet{karpinska2024thousandpairsnovelchallenge} and \citet{zhang2024inftybenchextendinglongcontext}, emphasize the model's ability to comprehend the global context. For example, \citet{karpinska2024thousandpairsnovelchallenge} manually curated a set of challenging questions based on various novels to evaluate global context understanding. It creates a realistic long-context benchmark emphasizing retrieval and global context understanding.
\subsection{Many-shot ICL with LCLMs}
As the context length of a model grows, the number of demonstrations that can be utilized in ICL also increases. Studies by \citet{li2024longcontextllmsstrugglelong}, \citet{bertsch2024incontextlearninglongcontextmodels}, and \citet{agarwal2024manyshotincontextlearning} have examined various properties of ICL under the many-shot setting. \citet{bertsch2024incontextlearninglongcontextmodels} explore whether models are merely performing retrieval tasks or genuinely understanding the tasks during many-shot ICL classification. Similarly, \citet{agarwal2024manyshotincontextlearning} analyzes the performance of tasks beyond classification in the many-shot context, using Gemini-Pro, and finds that additional demonstrations generally enhance task performance. Furthermore, \citet{li2024longcontextllmsstrugglelong} propose a long-context evaluation benchmark LongICLBench comprising many-shot ICL classification tasks, noting that current long-context models still face challenges in this area. 
None of the prior works has studied what skill each ICL task measures LCLMs for. LongICLBench mostly focuses on classification tasks, which may only evaluates the retrieval ability of LCLMs. 
Unlike previous studies, our work provides a more comprehensive analysis of many-shot ICL across a diverse set of tasks and multiple models. We introduce novel metrics to measure retrieval skills required for each task. We identify a set of ICL tasks suitable for evaluation and present a refined long-context evaluation benchmark with fine-grained categorization based on required similar-sample learning and all-sample learning skills. 
More related work on ICL can be found in Appendix \ref{app:related_work}. 

\section{Experiment Setting}
\label{experiment_setting}
\begin{table*}[t]
    \centering

\resizebox{0.8\linewidth}{!}{
    \begin{tabular}{lrrrrr}
    \toprule
    \textbf{Dataset}    & \textbf{Task Category}    & \textbf{Avg. Tokens / Shot} & \textbf{Max \# of Shots}  & \textbf{\# of Tasks}\\
    \midrule
    BANKING77 & Intent Classification & 13.13 & 5386 & 1 \\
    GoEmotions & Emotion Classification & 15.85 & 5480 &  1 \\
    DialogRE & Relation Classification & 233.27 & 395 &  1 \\
    TREC & Question Classification & 11.25 & 6272 &  1 \\
    CLINC150 & Intent Classification & 8.95 & 7252 &  1 \\
    MATH & Math reasoning & [185.52, 407.90] & [286, 653] &  4 \\
    GSM8K & Math reasoning & 55.78 & 784 &  1 \\
    BBH & Reasoning & [48.27, 243.01] & [406, 2660] &  4 \\
    GPQA & MQ - Science & [183.55, 367.02] & [314, 580] &  1 \\
    ARC & MQ - Science & [61.54, 61.54]  & [1997, 2301] &  2 \\
    XLSUM & New Summarization & 621.32 & 220 &  1 \\
    FLORES-200 & Translation & [63.63, 101.74] & [570, 1965] & 3 \\
    \bottomrule
    \end{tabular}
}
    \caption{Dataset information. GPT-4o tokenizer is used to calculate \# of tokens. Max \# of shots is the number of shots can be fitted into the 128k context window. For datasets that have multiple subtasks, we list the range for each value. We have 21 tasks in total.}
    \label{tbl:dataset}
\end{table*}


We select 11 open-weights models ranging from 3.8B to 123B parameters and one proprietary model. Our evaluation includes 12 datasets with 21 subtasks, spanning classification, summarization, reasoning, and translation domains. For each task, we randomly sample 200 data points from the test set, using the full test set if it contains fewer than 200 samples. 

For each task, we construct prompts for different context window sizes by incrementally adding new demonstrations from the training set to the prompt of the shorter context window size and duplicate training examples if they are insufficient to fill the context window. 
To ensure a fair comparison, we randomize the order of the demonstrations and consistently use the same set of examples across all context sizes. For simplicity, we apply greedy decoding across all models and conduct each experiment using three different random seeds. For prompt construction, we only include demonstrations and provide minimal task instruction. See Table \ref{tbl:example_prompts} for more details.
\subsection{Datasets}
Following previous works \citep{agarwal2024manyshotincontextlearning, li2024longcontextllmsstrugglelong, agarwal2024manyshotincontextlearning}, we include five datasets for \textbf{classification} tasks: BANKING77, GoEmotions, DialogRE, TREC, and CLINC150. For the \textbf{summarization} task, we use XLSUM, and for \textbf{translation}, we use FLORES-200. Additionally, we incorporate four datasets for \textbf{reasoning} tasks: MATH, BBH, and GPQA, and ARC. More details about each dataset can be found in Table \ref{tbl:dataset} and Appendix \ref{dataset}.

For the MATH, BBH, GPQA, and ARC tasks, we use accuracy as the evaluation metric. Macro F1-score is employed as the metric for all classification tasks. ROUGE-L \citep{lin-2004-rouge} is used for the XLSUM summarization task. ChrF \citep{popovic-2015-chrf} is applied for translation evaluation. 

\subsection{Models}
Below we list out the 12 models used in our experiment: Llama-3.1 8B and 70B \citep{dubey2024llama3herdmodels}, GLM-4-9B-Chat \citep{glm2024chatglmfamilylargelanguage}, Mistral Nemo (12B) and Large (123B) \citep{mistral_nemo_2024}, Qwen2 7B and 72B \citep{yang2024qwen2technicalreport}, Phi-3 mini (3.8B), small~(7B), and medium~(14B) \citep{abdin2024phi3technicalreporthighly}, and Jamba 1.5 Mini (12B/52B)~\citep{jambateam2024jamba15hybridtransformermambamodels}, and Gemini-1.5-Pro \citep{geminiteam2024gemini15unlockingmultimodal}.
We use the instruction-tuned version of all the models. For open-weights models with more than 50B, we run the quantized version of the models.\footnote{In Appendix \ref{quantized}, we show that quantized and unquantized models roughly exhibit the same trend.} More details about each model can be found in Appendix \ref{app:models}.

\section{Preliminary Study on many-shot ICL}
\begin{figure*}[h]
    \centering
    \begin{subfigure}{0.52\textwidth}
        \centering
        \includegraphics[width=\linewidth]{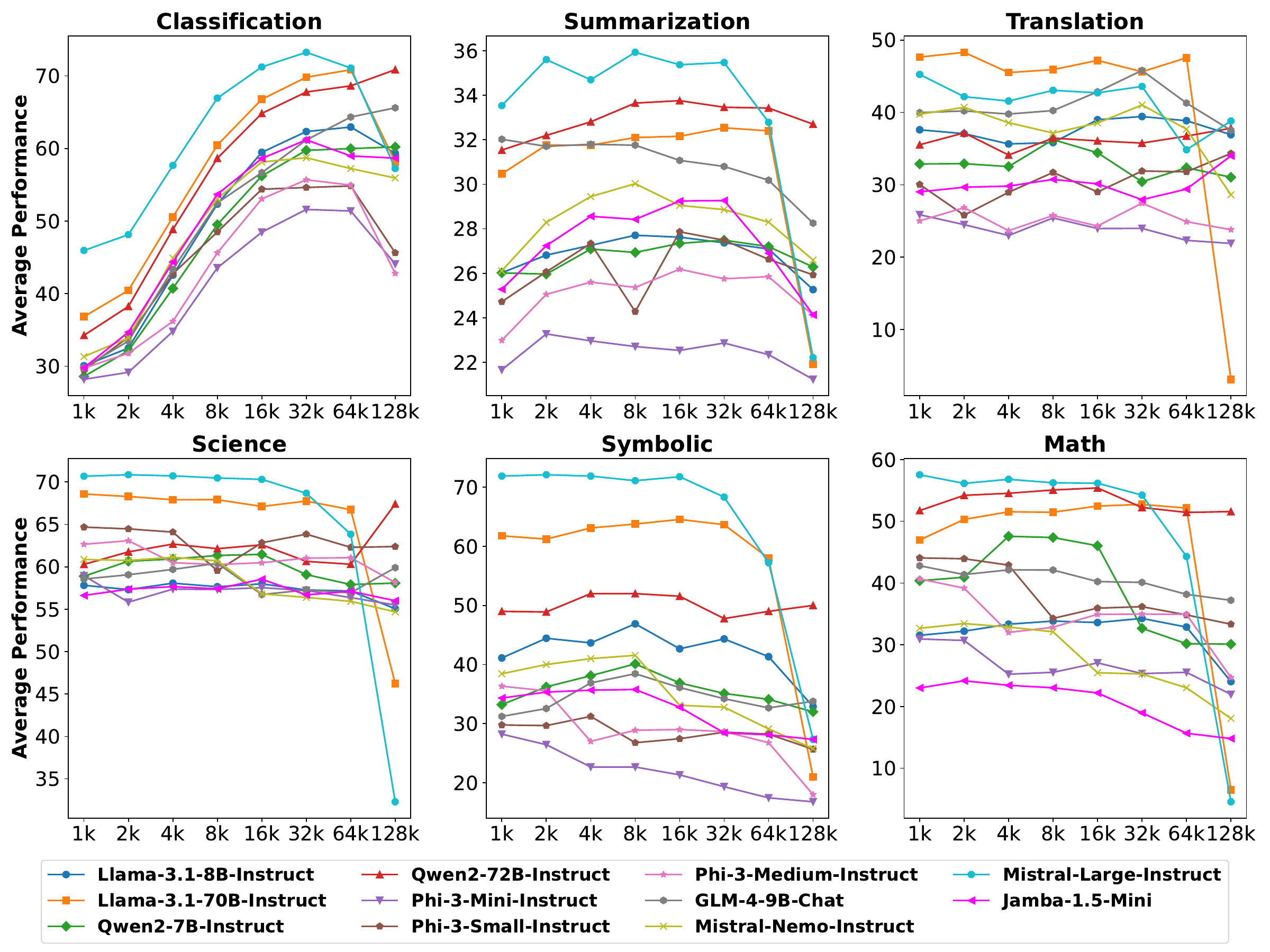}
        \subcaption{}
        \label{fig:categoriztion_perf}
    \end{subfigure}
    \hfill
    \begin{subfigure}{0.47\textwidth}
        \centering
        \includegraphics[width=\linewidth]{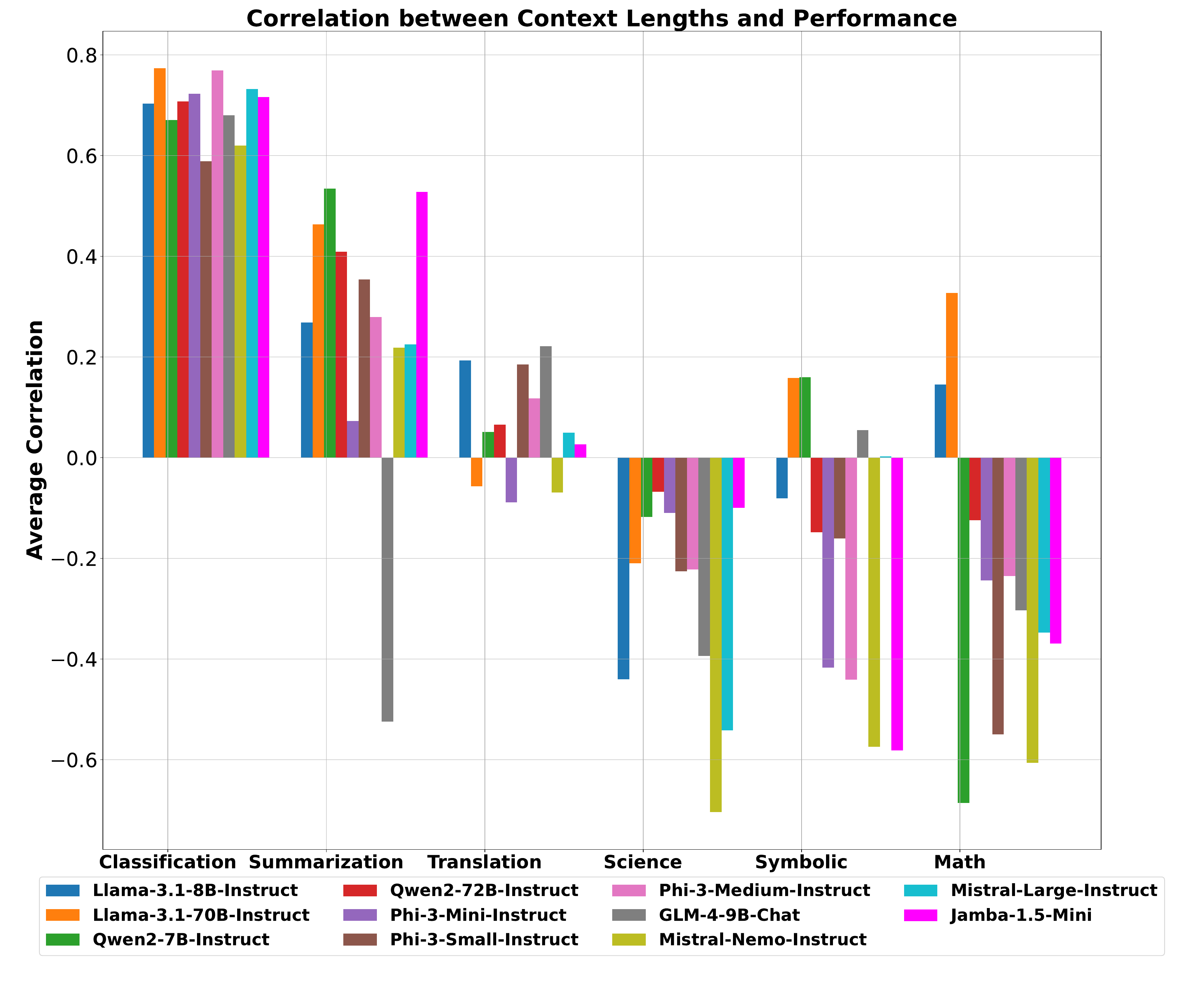}
        \subcaption{}
        \label{fig:correlation}
    \end{subfigure}
    \vspace{-8mm}
    \caption{
    (a) Aggregated performance of models over datasets in different categories of tasks. (b) Average pearson correlation coefficient between context lengths (1k to 64k) and the corresponding performance.}
    \label{fig:combined}
\end{figure*}

\vspace{-0.1cm}
In this section, we explore the extent to which many-shot ICL enhances models' performance across different task types. 
Previous work has either focused on only classification tasks \citep{bertsch2024incontextlearninglongcontextmodels} or studied only one specific model \citep{agarwal2024manyshotincontextlearning}.
In contrast, our analysis provides a comprehensive evaluation of many-shot ICL across both classification and generation tasks using eleven open-weights LCLMs. We collect tasks from previous work \citep{bertsch2024incontextlearninglongcontextmodels, agarwal2024manyshotincontextlearning, li2024longcontextllmsstrugglelong} from six categories: classification, translation, summarization, math reasoning, science reasoning, and symbolic reasoning.
The results, illustrated in Figure \ref{fig:combined}, include aggregated model performance across task types and the correlation coefficients between context lengths and performance from 1k to 64k. We also plot models' performance on individual tasks in Appendix \ref{task performance} and present more analysis on task categories in Appendix \ref{app:many_shot_ICL_analysis}.

\textbf{Classification performance steadily improves with more shots.} Figure \ref{fig:categoriztion_perf} demonstrates a consistent performance increase across all models as more demonstrations are added for classification tasks. This trend indicates a strong positive correlation between context length and performance, which is illustrated in Figure \ref{fig:correlation}. Given that some  classification tasks often involve extensive label spaces, e.g., CLINC150 has 150 classes, additional demonstrations provide models with exposure to more classes and thus enhance their ability to perform accurately. This is consistent with prior research findings \citep{bertsch2024incontextlearninglongcontextmodels}.

\textbf{Summarization shows gradual performance gains only.}  On summarization, most models exhibit a high correlation between context length and performance. However, there is a noticeable slowdown in the performance gains as the number of demonstrations increases. This suggests that while additional context may improve performance, it does so at a diminishing rate, particularly for smaller models like Llama-3.1-8B that struggle to leverage longer contexts effectively.

\textbf{Inconsistent trends in science and symbolic tasks.} For science and symbolic reasoning tasks, the performance trends are less predictable, with some models displaying minimal changes when seeing additional examples, while others benefit. \textit{This variability suggests that not all tasks lend themselves to the advantages of many-shot ICL equally}. 
Ideally, for every task, additional demonstrations should either improve performance or, at the very least, not harm it. A model with robust long-context capabilities should exhibit a non-decreasing performance trend as the context length increases. Given the inconsistent performance on non-classification tasks and even decreasing performance on some reasoning tasks, in the next section, we further investigate what aspects these datasets evaluate and identify a set of tasks useful for evaluating important skills of LCMLs. 

\section{Task Categorization: Similar-Sample Learning vs. All-Sample Learning}
\label{sec:Data Categorization}
To understand what skill each ICL task primarily requires from LCLMs, in this section, we measure the sample learning ratio of each task and divide them into \textit{similar-sample learning} vs. \textit{all-sample learning} tasks. We exclude translation tasks as tokenization can be inconsistent across languages and to avoid difference in multilingual capabilities affecting the results. Similarly, we omit GoEmotions because of its subjective nature. Given the resource constraint, we only choose a subset of models to measure the sample learning ratio. 

\subsection{Sample Learning Ratio}
\label{sec:retrieval load}
To identify similar-sample learning tasks, we propose a simple metric, \textbf{sample learning ratio (SLR)}, to assess whether tasks predominantly rely on models to retrieve relevant examples during many-shot ICL. 
Concretely, for each ICL task, we create two variants of the original demonstrations at each context size ranging from 1k to 64k by removing the 10\% most similar and the 10\% least similar examples. At context length $l$, the model's performance on these variants is then evaluated, and we have $\text{Perf}^{(l)}_{\text{most}}$ for removing similar examples and $\text{Perf}^{(l)}_{\text{least}}$ for removing dissimilar examples. Here we use BM25 retriever to calculate the similarity. We then average the ratios between $\text{Perf}^{(l)}_{\text{least}}$ and  $\text{Perf}^{(l)}_{\text{most}}$ for $l=1k$ to $l=64k$ as: 
\begin{figure*}[h]
    \centering
    \includegraphics[width=1\linewidth]{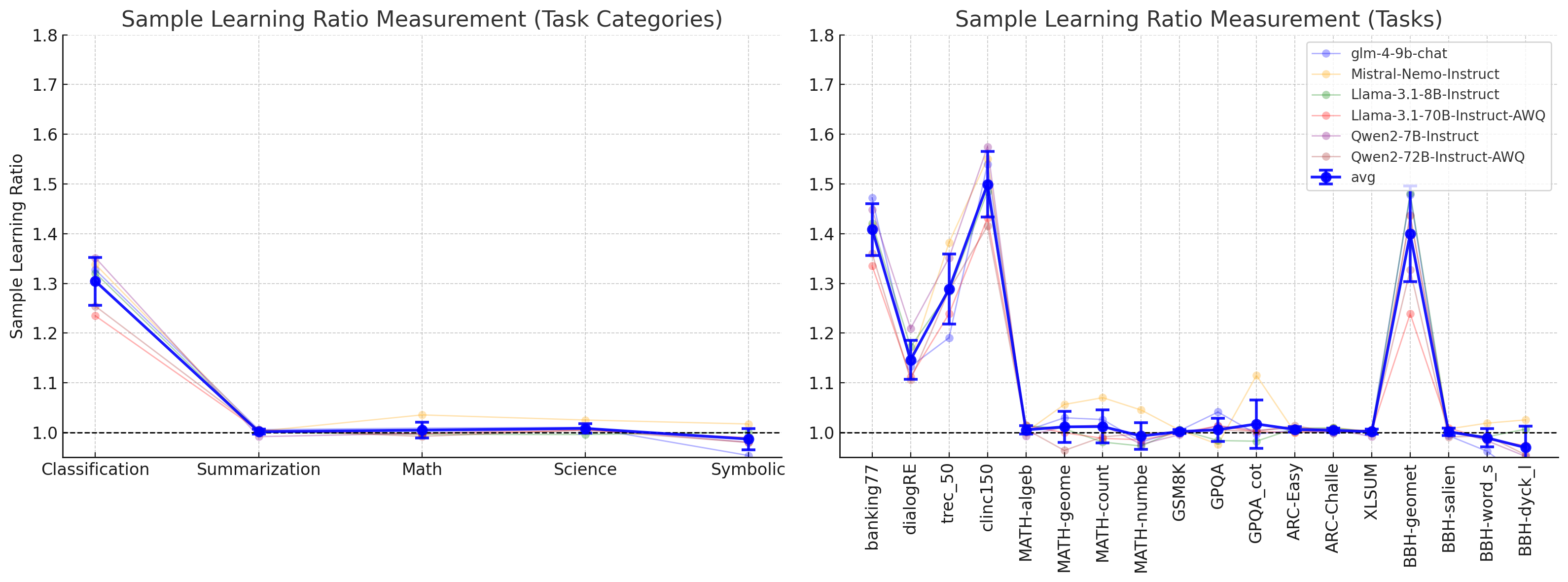}
    \vspace{-8mm}
    \caption{
    Sample Learning Ratio on different categories of tasks from 1k to 64k tokens. The ratio of 1 indicates models are not doing retrieval during ICL. Classification is the only category of tasks that has a very high ratio, which means classification tasks requires similar-sample retrieval during ICL. The rest of tasks is close to 1, and models' performance on these tasks does not rely on retrieving similar examples. 
    }
    \label{fig:retrieval_load}
\end{figure*}

\begin{equation}
    \label{retrieval_load_ratio_eq}
    \text{SLR} = \frac{1}{7}\sum_{l=1k}^{64k} \frac{\text{Perf}^{(l)}_{\text{least}}}{\text{Perf}^{(l)}_{\text{most}}}
\end{equation}

Intuitively, if a model predominantly relies on retrieval for a task, removing most similar examples will result in a more pronounced performance drop compared to removing dissimilar ones, which causes the ratio to be larger than 1. Conversely, if there is minimal difference between the two, it means the model does not retrieve similar examples to perform the task, and the ratio will be close to 1.

\paragraph{Classification tasks require retrieval of similar examples.} 
As shown in Figure \ref{fig:retrieval_load}, \textit{all classification tasks exhibit high SLR across the six models}. 
The BBH geometric shapes task also shows a high SLR, indicating that tasks like BANKING77, CLINC150, and TREC50 demand strong retrieval capabilities from the models. DialogRE has a relatively lower SLR, suggesting it requires moderate retrieval skills. Among the symbolic tasks, BBH-geometric\_shapes is the only reasoning task that has a high SLR. This task involves determining the geometric shape given a full SVG path element, making it similar to a classification task. The high SLR of classification tasks can possibly explain the largest positive correlation between performance and context lengths, as displayed in Figure \ref{fig:correlation}.

\paragraph{All-sample learning tasks.} In Figure \ref{fig:retrieval_load}, tasks such as the math problems and summarization, Dyck languages, translation error detection from BBH, and GPQA with explanations all have a low SLR. This means that \textit{they necessitate a greater degree of global context understanding rather than relying on the retrieval of relevant examples}. These tasks often involve complex reasoning challenges, for which models may lack pretraining skills to solve perfectly, underscoring the need for additional demonstrations or deeper task comprehension. 

\textbf{Additional analysis of SLR.} To ensure that performance loss is not caused by the absence of certain labels, we conduct an additional experiment in which the top k\% most similar examples are replaced with the most dissimilar examples that share the same labels. In addition to BM25, we also use the sentence transformer \citep{reimers-2019-sentence-bert} as the retriever. The result of this experiment exhibits the same trend: classification tasks show a higher SLR than non-classification tasks. More details can be found Appendix \ref{app:sample_learning_ratio_replacement}.

\section{ManyICLBench: A Many-shot ICL Benchmark to measure retrieval skill and global context understanding}
\definecolor{improvement_high}{RGB}{255, 51, 51}   
\definecolor{improvement_medium}{RGB}{255, 102, 102} 
\definecolor{improvement_low}{RGB}{255, 204, 204}   

\definecolor{drop_high}{RGB}{51, 153, 255}         
\definecolor{drop_medium}{RGB}{102, 178, 255}      
\definecolor{drop_low}{RGB}{204, 229, 255}         

\newcommand{\red}[1]{\colorbox{magenta!45}{#1}}
\newcommand{\blue}[1]{\colorbox{cyan!45}{#1}}
\newcommand{\shadecell}[2]{%
    \pgfmathsetmacro{\relchange}{(#2-#1)/#1*100} 
    \ifdim\relchange pt>0pt 
        \ifdim\relchange pt>100pt
            \cellcolor{improvement_high}#2
        \else
            \ifdim\relchange pt>50pt
                \cellcolor{improvement_medium}#2
            \else
                \cellcolor{improvement_low}#2
            \fi
        \fi
    \else 
        \ifdim\relchange pt<-20pt
            \cellcolor{drop_high}#2
        \else
            \ifdim\relchange pt<-10pt
                \cellcolor{drop_medium}#2
            \else
                \cellcolor{drop_low}#2
            \fi
        \fi
    \fi
}
\newcommand{\shade}[2]{%
    \pgfmathsetmacro{\relchange}{(#2-#1)/#1*100} 
    \ifdim\relchange pt>0pt 
        \ifdim\relchange pt>100pt
            \cellcolor{improvement_high}#2
        \else
            \ifdim\relchange pt>50pt
                \cellcolor{improvement_medium}#2
            \else
                \cellcolor{improvement_low}#2
            \fi
        \fi
    \else 
        \ifdim\relchange pt<-20pt
            \cellcolor{drop_high}#2
        \else
            \ifdim\relchange pt<-10pt
                \cellcolor{drop_medium}#2
            \else
                \cellcolor{drop_low}#2
            \fi
        \fi
    \fi
}

\begin{table*}[t]
    \centering

\resizebox{\linewidth}{!}{

    \begin{tabular}{lrrrrrrrrrr}
    \toprule
    \textbf{SSL Tasks} & \textbf{1k} & \textbf{2k} & \textbf{4k} & \textbf{8k} & \textbf{16k} & \textbf{32k} & \textbf{64k} & \textbf{128k} & \textbf{AVG.} & \textbf{AVG.L.}\\
    \midrule
    GLM-4-9b-Chat & 31.63 & \shadecell{31.63}{34.99} & \shadecell{31.63}{46.37} & \shadecell{31.63}{57.27} & \shadecell{31.63}{63.61} & \shadecell{31.63}{68.34} & \shadecell{31.63}{72.16} & \shadecell{31.63}{72.93} & 55.91 & 71.14 \\
    Mistral-Nemo-Instruct & 33.44 & \shadecell{33.44}{35.45} & \shadecell{33.44}{48.17} & \shadecell{33.44}{57.95} & \shadecell{33.44}{65.38} & \shadecell{33.44}{65.49} & \shadecell{33.44}{63.61} & \shadecell{33.44}{61.73} & 53.90 & 63.61 \\
    Mistral-Large-Instruct-AWQ & 49.15 & \shadecell{49.15}{51.23} & \shadecell{49.15}{60.78} & \shadecell{49.15}{71.95} & \shadecell{49.15}{77.10} & \shadecell{49.15}{79.45} & \shadecell{49.15}{77.77} & \shadecell{49.15}{61.89} & \textbf{66.16} & 73.04 \\
    Llama-3.1-8B-Instruct-AWQ & 32.13 & \shadecell{32.13}{34.63} & \shadecell{32.13}{45.76} & \shadecell{32.13}{57.39} & \shadecell{32.13}{66.18} & \shadecell{32.13}{70.02} & \shadecell{32.13}{70.55} & \shadecell{32.13}{65.85} & 55.31 & 68.81 \\
    Llama-3.1-70B-Instruct-AWQ & 38.75 & \shadecell{38.75}{42.87} & \shadecell{38.75}{53.98} & \shadecell{38.75}{66.07} & \shadecell{38.75}{73.12} & \shadecell{38.75}{76.56} & \shadecell{38.75}{78.48} & \shadecell{38.75}{65.56} & 61.92 & 73.53 \\
    Qwen2-7B-Instruct-AWQ & 30.18 & \shadecell{30.18}{34.03} & \shadecell{30.18}{44.40} & \shadecell{30.18}{54.85} & \shadecell{30.18}{62.92} & \shadecell{30.18}{65.91} & \shadecell{30.18}{66.94} & \shadecell{30.18}{66.38} & 53.20 & 66.41 \\
    Qwen2-72B-Instruct-AWQ & 36.41 & \shadecell{36.41}{41.89} & \shadecell{36.41}{54.24} & \shadecell{36.41}{65.33} & \shadecell{36.41}{73.39} & \shadecell{36.41}{76.53} & \shadecell{36.41}{77.51} & \shadecell{36.41}{77.47} & 62.85 & \textbf{77.17} \\
    Phi-3-Mini-Instruct & 30.27 & \shadecell{30.27}{30.90} & \shadecell{30.27}{38.09} & \shadecell{30.27}{48.14} & \shadecell{30.27}{53.58} & \shadecell{30.27}{57.29} & \shadecell{30.27}{56.83} & \shadecell{30.27}{48.72} & 45.48 & 54.28 \\
    Phi-3-Medium-Instruct & 31.73 & \shadecell{31.73}{33.55} & \shadecell{31.73}{39.10} & \shadecell{31.73}{49.83} & \shadecell{31.73}{58.29} & \shadecell{31.73}{61.17} & \shadecell{31.73}{60.63} & \shadecell{31.73}{45.32} & 47.45 & 55.70 \\
    Phi-3-Small-Instruct & 31.48 & \shadecell{31.48}{36.27} & \shadecell{31.48}{46.20} & \shadecell{31.48}{54.34} & \shadecell{31.48}{59.63} & \shadecell{31.48}{59.73} & \shadecell{31.48}{60.20} & \shadecell{31.48}{48.97} & 49.60 & 56.30 \\
    Jamba-1.5-Mini & 32.10 & \shadecell{32.10}{36.91} & \shadecell{32.10}{48.61} & \shadecell{32.10}{60.29} & \shadecell{32.10}{66.05} & \shadecell{32.10}{68.33} & \shadecell{32.10}{66.02} & \shadecell{32.10}{65.17} & 55.44 & 66.51 \\
    Gemini-1.5-Pro & 36.40 & \shadecell{36.40}{47.31} & \shadecell{36.40}{58.01} & \shadecell{36.40}{65.49} & \shadecell{36.40}{71.43} & \shadecell{36.40}{74.22} & \shadecell{36.40}{72.43} & \shadecell{36.40}{72.42} & 62.21 & 73.03 \\
    \midrule
    \textbf{ASL Tasks} & \textbf{1k} & \textbf{2k} & \textbf{4k} & \textbf{8k} & \textbf{16k} & \textbf{32k} & \textbf{64k} & \textbf{128k} & \textbf{AVG.} & \textbf{AVG.L.} \\
    \midrule
    GLM-4-9b-Chat & 40.51 & \shadecell{40.51}{40.28} & \shadecell{40.51}{42.04} & \shadecell{40.51}{42.78} & \shadecell{40.51}{40.70} & \shadecell{40.51}{40.46} & \shadecell{40.51}{38.85} & \shadecell{40.51}{39.13} & 40.59 & 39.48 \\
     Mistral-Nemo-Instruct & 38.25 & \shadecell{38.25}{39.07} & \shadecell{38.25}{39.28} & \shadecell{38.25}{38.99} & \shadecell{38.25}{33.06} & \shadecell{38.25}{32.83} & \shadecell{38.25}{30.46} & \shadecell{38.25}{27.11} & 34.88 & 30.13 \\
Mistral-Large-Instruct-AWQ & 61.47 & \shadecell{61.47}{61.10} & \shadecell{61.47}{61.23} & \shadecell{61.47}{60.87} & \shadecell{61.47}{60.86} & \shadecell{61.47}{58.84} & \shadecell{61.47}{50.01} & \shadecell{61.47}{16.69} & 53.88 & 41.85 \\
     Llama-3.1-8B-Instruct & 37.31 & \shadecell{37.31}{38.84} & \shadecell{37.31}{41.25} & \shadecell{37.31}{40.79} & \shadecell{37.31}{39.83} & \shadecell{37.31}{39.77} & \shadecell{37.31}{39.12} & \shadecell{37.31}{34.41} & 38.92 & 37.77 \\
Llama-3.1-70B-Instruct-AWQ & 53.32 & \shadecell{53.32}{54.84} & \shadecell{53.32}{55.76} & \shadecell{53.32}{55.87} & \shadecell{53.32}{56.42} & \shadecell{53.32}{56.34} & \shadecell{53.32}{54.42} & \shadecell{53.32}{18.58} & 50.69 & 43.12 \\
         Qwen2-7B-Instruct & 39.52 & \shadecell{39.52}{41.96} & \shadecell{39.52}{45.17} & \shadecell{39.52}{45.39} & \shadecell{39.52}{45.50} & \shadecell{39.52}{37.29} & \shadecell{39.52}{36.97} & \shadecell{39.52}{33.99} & 40.72 & 36.09 \\
    Qwen2-72B-Instruct-AWQ & 48.01 & \shadecell{48.01}{49.24} & \shadecell{48.01}{50.32} & \shadecell{48.01}{50.70} & \shadecell{48.01}{50.97} & \shadecell{48.01}{48.20} & \shadecell{48.01}{47.98} & \shadecell{48.01}{48.16} & 49.20 & 48.11 \\
       Phi-3-Mini-Instruct & 33.54 & \shadecell{33.54}{32.97} & \shadecell{33.54}{29.80} & \shadecell{33.54}{29.75} & \shadecell{33.54}{30.12} & \shadecell{33.54}{28.78} & \shadecell{33.54}{28.06} & \shadecell{33.54}{25.76} & 29.85 & 27.53 \\
     Phi-3-Medium-Instruct & 41.59 & \shadecell{41.59}{40.91} & \shadecell{41.59}{34.85} & \shadecell{41.59}{35.63} & \shadecell{41.59}{36.91} & \shadecell{41.59}{36.84} & \shadecell{41.59}{36.38} & \shadecell{41.59}{28.31} & 36.43 & 33.84 \\
      Phi-3-Small-Instruct & 41.61 & \shadecell{41.61}{41.61} & \shadecell{41.61}{41.61} & \shadecell{41.61}{35.58} & \shadecell{41.61}{37.17} & \shadecell{41.61}{37.73} & \shadecell{41.61}{36.91} & \shadecell{41.61}{35.33} & 38.44 & 36.65 \\
            Jamba-1.5-Mini & 31.96 & \shadecell{31.96}{33.08} & \shadecell{31.96}{32.97} & \shadecell{31.96}{32.70} & \shadecell{31.96}{31.66} & \shadecell{31.96}{28.82} & \shadecell{31.96}{27.14} & \shadecell{31.96}{25.87} & 30.53 & 27.28 \\
    Gemini-1.5-Pro & 57.87 & \shadecell{57.87}{63.39} & \shadecell{57.87}{64.15} & \shadecell{57.87}{66.78} & \shadecell{57.87}{68.02} & \shadecell{57.87}{67.78} & \shadecell{57.87}{66.14} & \shadecell{57.87}{66.42} & \textbf{65.07} & \textbf{66.78} \\
\hline
    \bottomrule
    \end{tabular}
}
    \caption{
    Model performance on SSL and ASL tasks. AVG. is the average model performance of all context lengths. AVG.L. is the average model performance of 32k, 64k and 128k. \red{Red} indicates performance improvement compared to 1k.  \blue{Blue} indicates performance downgrade compared to 1k. A darker color means higher improvement or downgrade. \textbf{BOLD} number means the largest number of a column. Many models start downgrading their performance after 32k on SSL tasks. On global context ASL tasks, many models start struggling even before 16k.}
    \label{tbl:retrieval_global}
\end{table*}

In this section, we present a new long-context benchmark \data, designed to evaluate LCLMs' retrieval skills and global context understanding capabilities using the ICL setup. 
Based on the results from Section \ref{sec:Data Categorization}, we group tasks into two types:

$\bullet$ \textbf{5 SSL Tasks}: BANKING77, dialogRE, TREC50, CLINC150, and the geometric shape task from BBH. 

$\bullet$ \textbf{11 ASL Tasks}: all math tasks, summarization task, GPQA with explanations, ARC\_challenge, and all BBH tasks except geometric shapes.

We exclude ARC\_Easy and GPQA since ARC\_Challenge and GPQA\_CoT are already included in the benchmark. Evaluation results of popular LCLMs are summarized in Table \ref{tbl:retrieval_global}. 

\paragraph{Most models struggle at retrieving examples after 32k length.} Up to a context length of 16k, \textit{all models demonstrate a steady performance increase, indicating effective retrieval from shorter contexts}. However, performance begins to decline after reaching 32k tokens, particularly for the Mistral family and Jamba models. 
After 64k, the Llama 3.1 family and the mini and medium versions of Phi-3 exhibit a notable downgrade in performance. In contrast, the Qwen-2 family maintains robust performance, with minimal degradation from 64k to 128k. Remarkably, only GLM-4 continues to improve in retrieval performance beyond 64k, indicating its impressive retrieval capabilities within a very long context window, while larger models such as Mistral-Large and Llama-3.1-70B exhibit the most significant performance losses as context length increases, suggesting that size alone does not ensure superior long-context retrieval ability.

\paragraph{Challenges in ASL tasks.} ASL tasks prove to be more challenging, with \textit{many models struggling even at short context lengths like 2k or 4k}. Only the Llama 3.1 family, Qwen2 family, GLM-4, and Gemini models effectively leverage many demonstrations up to 16k. At 32k, only the Llama 3.1 and Gemini models sustain performance. As context length extends from 32k to 128k, all models experience performance degradation, highlighting that current architectures still struggle to grasp global context and utilize demonstrations effectively. Notably, Qwen2-72B, GLM-4, and Gemini are the only models that do not experience significant performance drops in this category. 

\paragraph{The paradox of model size.} Despite the common assumption that larger models possess greater capabilities, our findings illustrate that larger models can experience more substantial performance losses compared to smaller models if not trained adequately on long-context data. For instance, Mistral-Large (123B) shows optimal performance from 1k to 32k but experiences a dramatic drop beyond 32k, which is worse than Phi-3-Mini (3.8B). A similar trend is observed with Llama-3.1-70B at 128k. Both underscore the importance of targeted training for long-context tasks. Although the large models we test are quantized, we believe the trend will hold even for the full-precision models.  

\paragraph{Llama 3.1 performance and training limitations.} The Llama 3.1 models initially capitalize on additional demonstrations effectively up to 64k but suffer significant performance declines at 128k. This pattern aligns with trends observed in other long-context evaluation benchmarks \citep{hsieh2024rulerwhatsrealcontext}. We suspect that these performance drops are linked to insufficient training with long-context data during the supervised fine-tuning (SFT) stage. According to \citet{dubey2024llama3herdmodels}, the average token count for long-context datasets is around 38k, indicating limited exposure for models to effectively learn from data points at 128k lengths.

\paragraph{Gemini exhibits robustness.} Similar to other open-weight models, Gemini-1.5-Pro shows performance degradation on SSL tasks beyond 32k tokens. However, it uniquely maintains strong retrieval capabilities and stable performance up to 128k tokens, alongside models like Qwen-2-72B and GLM-Chat-9B. On ASL tasks, Gemini-1.5-Pro significantly surpasses other open-weight models, demonstrating exceptional proficiency in utilizing the global context effectively and leveraging all provided demonstrations.

\paragraph{Additional analysis on all-sample learning.} To investigate whether ASL tasks genuinely benefit from additional demonstrations and to assess the models' ability to understand the global context, we compare their performance using unique versus duplicated demonstrations. Our analysis reveals that some ASL tasks do not benefit significantly from additional unique demonstrations, suggesting that models do not always fully leverage all available demonstrations during ICL. Further detailed analyses are provided in Appendix \ref{app:global context understanding}.

\begin{figure}[h]
    \centering
    \includegraphics[width=1\linewidth]{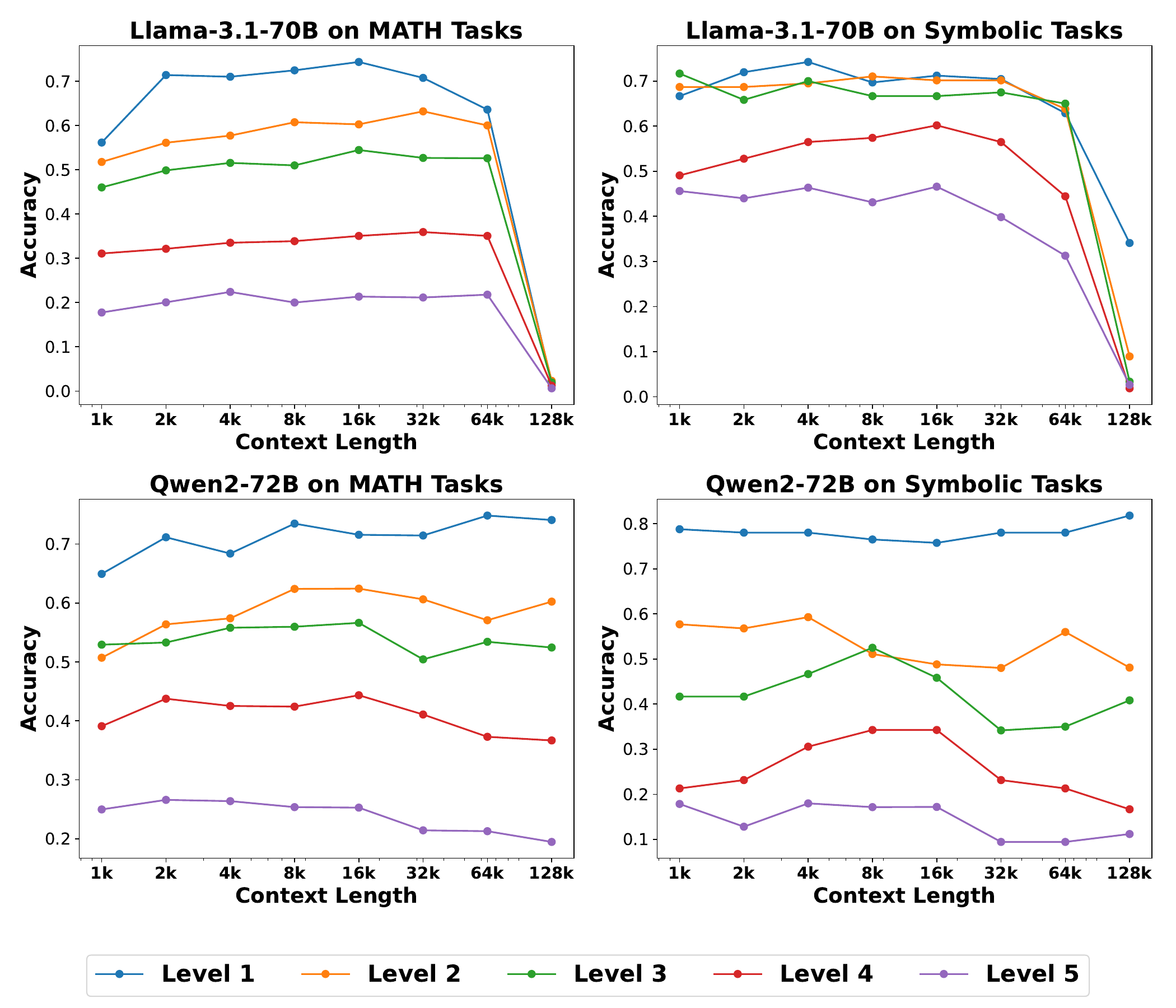}
    \caption{
    Llama-3.1-70B-Instruct and Qwen2-72B-Instruct aggregated performance on four MATH tasks and two symbolic tasks from Section \ref{error_analysis}. For the MATH tasks, we use the dataset's difficulty labels. For the symbolic tasks, we approximate difficulty based on word or symbol length.
    }
    \label{fig:aggregated_difficulty_performance}
\end{figure}
\label{error_analysis}

\section{Error Analysis}

\subsection{Why LCLMs fail on ASL tasks?}
To better understand how LCLMs fail on ASL tasks, we analyze the performance of Llama-3.1-70B-Instruct and compare it with Qwen-2-72B-Instruct. We focus on non-multiple-choice tasks, including four subtasks from MATH, two symbolic tasks from BBH (word\_sorting and dyck\_languages), and XLSUM. All the error analysis examples can be found in Appendix~\ref{app:error_analysis}.

\paragraph{LCLMs' reasoning ability downgrades at long context lengths.}
As shown in Figure~\ref{fig:aggregated_difficulty_performance}, when the context length increases, Llama-3.1-70B-Instruct tends to lose performance more significantly than Qwen-2-72B-Instruct. Across both easier and more difficult tasks, Qwen-2-72B-Instruct maintains relatively stable performance. A notable observation is that for both models, more difficult tasks degrade at an earlier context length, while easier tasks show a drop only at a later point. For instance, on symbolic tasks, Llama-3.1-70B-Instruct starts losing accuracy on level-4 and level-5 tasks before the simpler level-1 task. 

\paragraph{Llama-3.1-70B-Instruct loses reasoning capability at 128k length.}
On MATH tasks at 128k tokens, Llama-3.1-70B-Instruct often produces repetitive reasoning and gets stuck in loops. Its chain of thought becomes oversimplified, frequently relying on basic sentence structures. In Table~\ref{tbl:math_algebra}, the solution at 128k tokens contains many repeated segments, whereas Qwen-2-72B-Instruct still generates valid, consistent reasoning. Furthermore, Llama-3.1-70B-Instruct’s mathematical accuracy suffers: in Table \ref{tbl:math_counting}, although it correctly calculates \(\binom{9}{2} = 36\) at shorter lengths, it incorrectly outputs 42 at 128k tokens.

On symbolic tasks, as test examples become more complex with additional symbols and words, Llama-3.1-70B-Instruct fails to solve problems it could handle at shorter contexts. While Qwen-2-72B-Instruct still performs well on easier examples at 128k tokens, Llama-3.1-70B-Instruct shows a noticeable drop.

\paragraph{Weaker instruction-following at 128k length.}
For summarization tasks, the summaries generated by Llama-3.1-70B-Instruct at 128k tokens become longer and often contain details not present in the reference. In Table~\ref{tbl:xlsum_example}, Qwen-2-72B-Instruct maintains concise, one-sentence outputs at both 16k and 128k tokens, but Llama-3.1-70B-Instruct adds extraneous information and deviates from the requested one-sentence format. Figure~\ref{fig:XLSUM_length} also illustrates a significant increase in output length for Llama-3.1-70B-Instruct at 128k tokens.

\paragraph{Overly long generation at 128k length.}
Figure~\ref{fig:MATH_length} shows that Llama-3.1-70B-Instruct’s chain-of-thought for MATH tasks grows excessively at 128k tokens, partly due to repetitive loops. This trend indicates that beyond a certain context length, the model’s reasoning becomes unfocused and verbose.

\begin{figure}[h]
    \centering
    \includegraphics[width=1\linewidth]{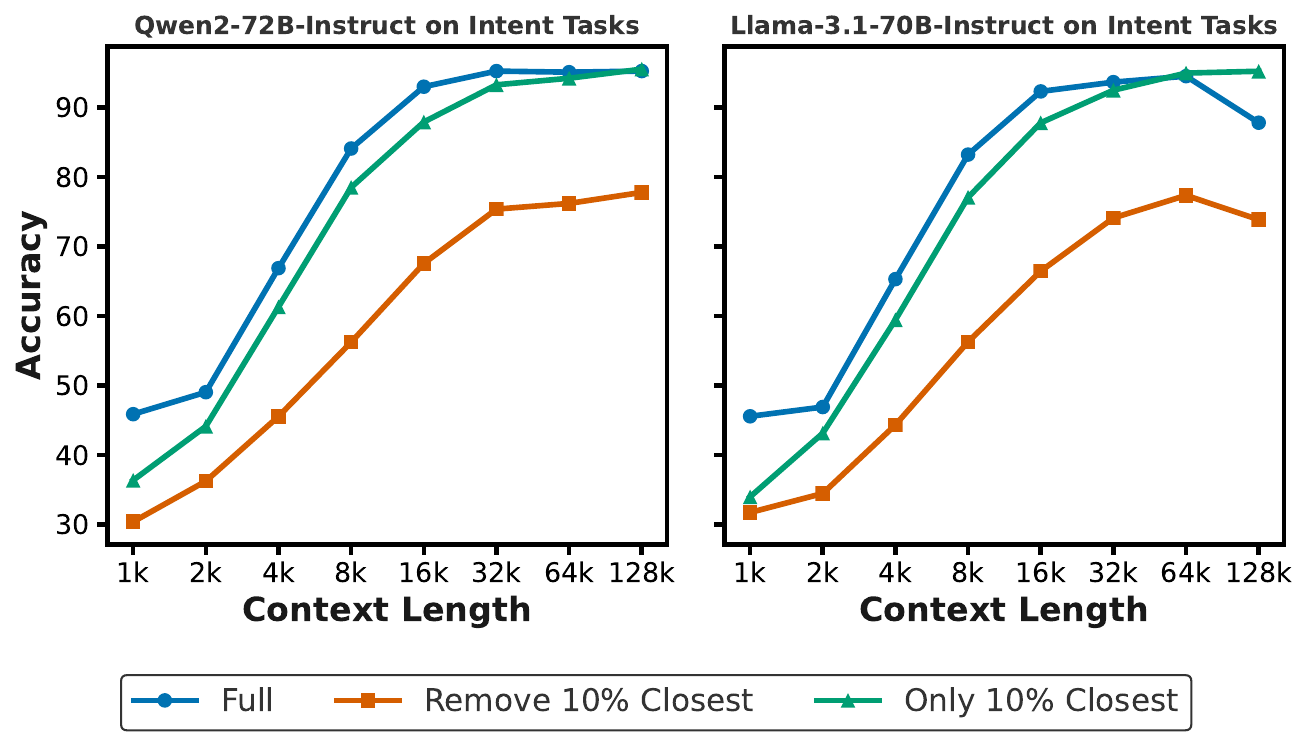}
    \caption{
    Aggregated performance of Llama-3.1-70B-Instruct and Qwen2-72B-Instruct on Banking77 and CLINC150 intent classification tasks. \textbf{Full}: All demonstrations are used during evaluation. \textbf{Remove 10\% closest}: The 10\% most similar demonstrations to each test instance are removed from the full set. \textbf{Only 10\% closest}: Only the 10\% most similar demonstrations are used.
    }
    \label{fig:intent_tasks}
\end{figure}
\label{intent_tasks}

\subsection{Why do LCLMs Perform Better on SSL Tasks?}

To better understand why LCLMs excel at SSL tasks, we closely examine the performance of the same two models, Qwen2-72B-Instruct and Llama-3.1-70B-Instruct, on two representative intent classification tasks, BANKING77 and CLINC150. We compare three distinct demonstration conditions: (1) utilizing all available demonstrations (full), (2) removing the 10\% most similar demonstrations from the full set, and (3) using only the 10\% most similar demonstrations to each test instance.

As illustrated in Figure \ref{fig:intent_tasks}, both models achieve near-identical performance when using only the 10\% most similar demonstrations compared to the full set of demonstrations, particularly as the context length increases. Notably, performance converges at the 64k token context length, suggesting that a small subset of relevant examples is sufficient for strong accuracy on these SSL tasks. Conversely, excluding the most relevant 10\% of demonstrations significantly degrades the performance of both models across all context lengths. This emphasizes that SSL tasks inherently depend on effective retrieval and utilization of a small, highly relevant subset of examples.

This behavior aligns with findings in recent literature \citep{bertsch2024incontextlearninglongcontextmodels}, where retrieving as little as 5\% of the most similar demonstrations can match or exceed the effectiveness of using randomly sampled examples. Thus, even at extensive context lengths (e.g., 128k tokens), models primarily benefit from a focused subset of demonstrations, highlighting the retrieval-dominated nature of SSL tasks. In contrast, ASL tasks offer no retrieval shortcuts, compelling models to comprehensively integrate information from all provided demonstrations to solve the tasks.

\section{Conclusion}

We investigated many-shot in-context learning (ICL) across various tasks using different open-weight models, assessing their suitability for evaluating long-context language models (LCLMs). Our findings indicate that classification and summarization tasks consistently benefit from additional demonstrations, while other tasks do not. 
To identify a set of tasks suitable for long-context evaluation, we introduced the concept of sample learning ratio (SLR) to assess the retrieval demands of different tasks. This analysis revealed that classification tasks predominantly rely on the model’s retrieval capabilities. 
Based on these insights, we categorized tasks into two distinct groups: similar-sample learning (SSL) tasks and all-sample learning (ASL) tasks. Furthermore, we introduced a novel many-shot ICL benchmark, \textbf{ManyICLBench}, designed to evaluate both retrieval and global context understanding skills of LCLMs.
Benchmarking open-weight LCLMs on ManyICLBench revealed that most models struggle with ASL tasks at lengths beyond 16k tokens. In contrast, performance on SSL tasks tends to decline after 32k tokens. 

\section*{Acknowledgments}
This work is supported in part by National Science Foundation through grant 2046016. We are grateful to ICLR and ARR reviewers for their thoughtful comments. 

\section*{Limitation}
Our study focuses solely on the in-context learning setup and does not address other setups where long-context abilities are important, such as retrieval-augmented generation. Additionally, we restrict our evaluation and benchmark to English ICL tasks, as most existing long-context benchmarks are English-centric. Expanding the benchmark to include multilingual tasks is an important direction for future work.

Moreover, we acknowledge that data contamination is a crucial area to investigate; however, it falls beyond the scope of our current study. We argue that our conclusions remain valid irrespective of whether the model has previously encountered the data, as additional demonstrations should ideally enhance, rather than hinder, performance. Any observed performance degradation is likely attributable to the model’s limitations in effectively handling long-context content. A detailed analysis of the effects of data contamination is an important avenue for future research.

Lastly, given the rapid development and iteration of LCLMs, our findings may not fully generalize to newer models that may be trained to handle long context differently from existing ones. Nevertheless, we believe our findings are useful for future efforts toward building models with strong long-context understanding.
\bibliography{custom}

\clearpage
\appendix
\section{Additional Related Work}
\label{app:related_work}
In-context learning (ICL) enables models to quickly recognize and perform tasks during inference by conditioning on a set of provided demonstrations \citep{brown2020languagemodelsfewshotlearners}. Many previous works have sought to understand the mechanisms behind in-context learning (ICL). \citet{xie2022explanationincontextlearningimplicit} suggests that models implicitly perform Bayesian inference during inference, retrieving relevant skills learned during pretraining. Additionally, \citet{lin2024dualoperatingmodesincontext} introduces the concept of a dual operating mode in ICL: task learning and task retrieval. With sufficient demonstrations, models can adapt to unseen tasks learned during pretraining, thereby enhancing performance as the number of demonstrations increases. To explore how many-shot ICL operates, \citet{bertsch2024incontextlearninglongcontextmodels} modified the attention patterns by restricting attention among individual examples. Their findings suggest that performance improvements primarily arise from retrieving similar examples rather than comprehending the task. However, their experiment is limited to classification tasks. It may also be biased when comparing full attention and block attention, as block attention allows access to more demonstrations. Our work tries to design better experiments to investigate during many-shot ICL what skill each task mainly requires from LCLMs.  
\section{Datasets}
\label{dataset}

\textbf{BANKING77} \citep{casanueva2020efficientintentdetectiondual} is an intent classification task in the banking domain. It has over 10k customer service queries labeled with 77 intents.

\textbf{GoEmotions} \citep{demszky2020goemotionsdatasetfinegrainedemotions} contains 58 Reddit comments labeled for 27 emotion categories or Neutral.

\textbf{DialogRE} \citep{yu-etal-2020-dialogue} is a relation extraction dataset that is built based on transcripts of an American TV show Friends. It comprises 10,168 relation triples for 1,788 dialogues and 36 total relations types. We only focus on relation classification for this dataset.

\textbf{TREC} \citep{li-roth-2002-learning, hovy-etal-2001-toward} is a question classification dataset with six coarse and 50 fine class labels. It contains 5,500 questions in the training set and 500 in the test set.

\textbf{CLINC150} \citep{larson-etal-2019-evaluation} is an intent classification dataset with 150 intents from 10 domains. 

\textbf{MATH} \citep{hendrycksmath2021} is a dataset of 12,5000 challenging completion mathematics problems. Each problem has a full step-by-step solution. We use four subdomains from the dataset: algebra, geometry, counting and probability, and number theory.

\textbf{GSM8K} \citep{hendrycksmath2021} consists of 8.5K high quality grade school math problems created by human problem writers. These problems take between 2 and 8 steps to solve, and solutions primarily involve performing a sequence of elementary calculations using basic arithmetic operations (+ - / *) to reach the final answer. 

\textbf{BBH} \citep{srivastava2022beyond} is a subset of 23 challenging BIG-Bench tasks \citep{suzgun2022challenging}, which include task categories such as mathematics, commonsense reasoning, and question answering. We use four subtasks from BBH-Hard: geometric shape, salient translation error detection, word sorting, and dyck languages.

\textbf{ARC} \citep{allenai:arc} is a dataset of 7,787 genuine grade-school level, multiple-choice science questions. The dataset is partitioned into a Challenge Set and Easy Set, where the former contains only questions answered incorrectly by both a retrieval-based algorithm and a word co-occurrence algorithm.

\textbf{GPQA} \citep{rein2023gpqa} is a dataset of 448 multiple-choice questions with detailed explanations written by domain experts in biology, physics, and chemistry.

\textbf{XLSUM} \citep{hasan-etal-2021-xl} is a summarization dataset that focuses on news articles from BBC. In this work, we focus only on English news articles.

\textbf{FLORES-200} \citep{nllb2022} is a translation benchmark that contains many low-resource languages. We follow \citet{agarwal2024manyshotincontextlearning} and choose the translation task from Tamil to English. Additionally, we also test models on Chinese and Spanish.

\section{Models}
\label{app:models}
\textbf{Llama-3.1 8B and 70B}~\citep{dubey2024llama3herdmodels}: We use both the 8B and 70B Llama 3.1 Instruction models. These multilingual models are trained on a 128k context window using position interpolation. The models are further fine-tuned with synthetic long-text Supervised Fine-Tuning (SFT) data and also undergo Direct Preference Optimization (DPO) \citep{rafailov2024directpreferenceoptimizationlanguage}.

\textbf{GLM-4-9B-Chat} \citep{glm2024chatglmfamilylargelanguage}: This is a 9-billion-parameter multilingual model, also trained on a 128k context window with position interpolation. It is further fine-tuned with labeled long-text SFT data and undergoes a DPO stage.

\textbf{Mistral Family} \citep{mistral_nemo_2024}: We use both 12-billion-parameter and 123-billion-parameter multilingual models, trained on a 128k context window.

\textbf{Qwen2 7B and 72B} \citep{yang2024qwen2technicalreport}: These two models are trained with a context size of 32k tokens, and their context window is extended to 128k by YARN \citep{peng2023yarnefficientcontextwindow}, a dynamic position interpolation technique. 

\textbf{Phi-3}~\citep{abdin2024phi3technicalreporthighly}: We use the mini (3.8B), small (7B), and medium (14B) versions of Phi-3 models. They are trained with the context size of 4k tokens on high quality data, and LongRope \citep{ding2024longropeextendingllmcontext} extends their context size to 128k. 

\textbf{Jamba-1.5-Mini}~\citep{jambateam2024jamba15hybridtransformermambamodels}: It's a hybrid SSM-Transformer model with 12B of active parameters and 52B of total parameters with a context size of 256k tokens. 

\textbf{Gemini-1.5-Pro} \citep{geminiteam2024gemini15unlockingmultimodal}: It is a commercial model introduced by Google and has a context size of 2 million tokens.

\section{Quantization vs. Regular}
\label{quantized}
\begin{figure}[h]
    \centering
    \includegraphics[width=1\linewidth]{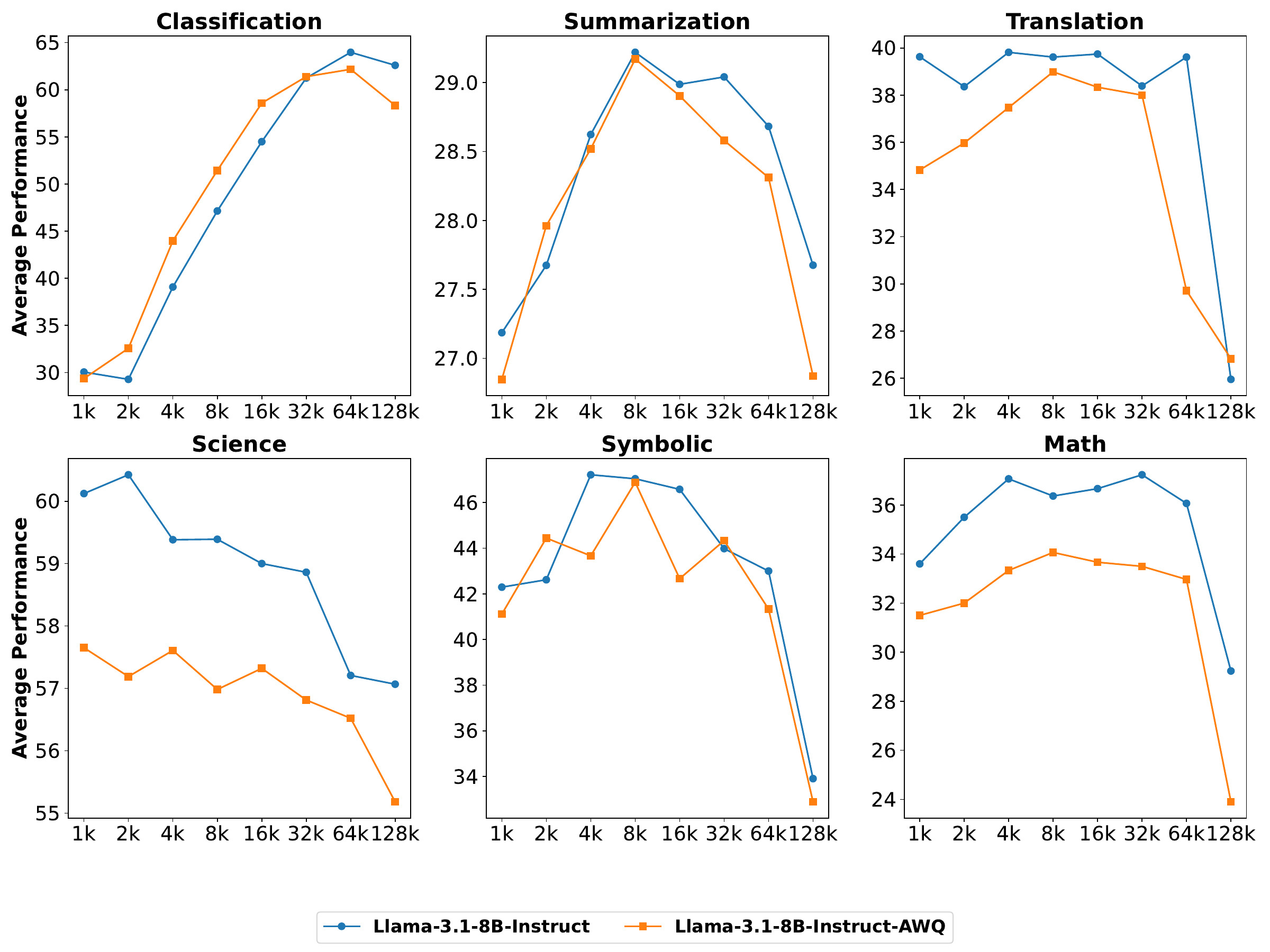}
    \caption{
    Comparison between Llama-3.1-8B and 4-bit quantized Llama-3.1-8B. There are some performance gaps between two models on translation, science, and math tasks, but with the increasing context size, the performance trend is the same for both models.
    }
    \label{fig:Llama-3.1-8B-quant}
\end{figure}

\begin{figure}[h]
    \centering
    \includegraphics[width=1\linewidth]{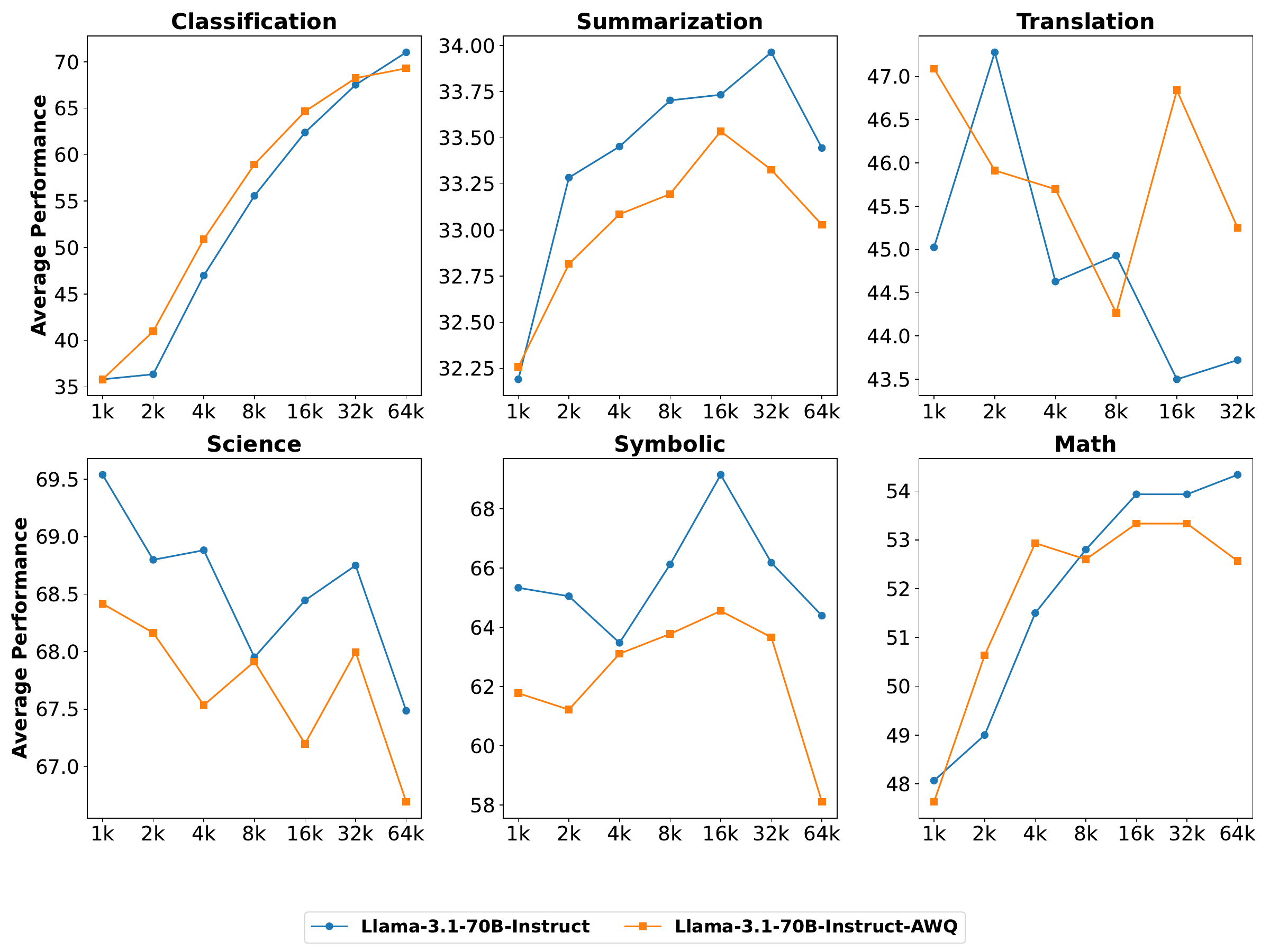}
    \caption{
    Comparison between Llama-3.1-70B and 4-bit quantized Llama-3.1-70B. Similar to the smaller model, the performance trends hold for both models except the translation tasks. In our benchmark, we exclude all the translation tasks because of the inconsistent multilingual ability of LCLMs. 
    }
    \label{fig:Llama-3.1-70B-quant}
\end{figure}

We compare the 4-bit quantized version and unquantized version of both Llama-3.1 8B and Llama-3.1-70B. In both Figure \ref{fig:Llama-3.1-8B-quant} and Figure \ref{fig:Llama-3.1-70B-quant}, we can observe that the quantized version experiences a little performance drop but exhibits the same trend as the unquantized version with the increasing context length. 

\section{More analysis on Section 4 many-shot ICL}
\label{app:many_shot_ICL_analysis}
\textbf{Subjective tasks do not benefit from more examples.}  The GoEmotions task, though being a classification problem, exhibits a fluctuating performance trend across all models with increasing shots in Figure \ref{fig:classification}. We attribute this inconsistency to the subjective nature of the task, where nuanced emotional categories may lead to low annotator agreement \citep{demszky2020goemotionsdatasetfinegrainedemotions}. This variance in the annotated labels may results in a weaker correlation between context length and performance. 
This finding highlights a limitation in using ICL tasks with ambiguous ground truths to evaluate LCLMs, as their performance does not improve with more demonstrations.

\textbf{Models' performance fluctuates on translation tasks.} As shown in Figure \ref{fig:translation}, the performance curves for all models across different languages differ. For the low-resource language, models show larger performance gap than those in the high-resource language, e.g., Spanish. In Chinese, models become spikier than in other languages across different context sizes. In Figure \ref{fig:categoriztion_perf}, translation tasks show a very flat curve, with no significant improvement as the number of demonstrations increases. This result contrasts with \citet{agarwal2024manyshotincontextlearning}, where the Gemini-1.5 Pro model demonstrated consistent performance improvements in Kurdish and Tamil translation tasks as the context size increased. We think the performance inconsistency is caused by the mismatched multilingual capability of models and different model sizes.

\textbf{Math tasks benefit from additional demonstrations, particularly for stronger models.} In math reasoning tasks, only the Llama-3.1 and Qwen2 model families show significant performance improvements with additional demonstrations. Notably, Qwen2 performance plateaus at 16k length, while Llama-3.1 continues to improve until 64k. The models with larger parameter sizes tend to exhibit more consistent performance gains, supporting findings from \citet{agarwal2024manyshotincontextlearning} who have demonstrated that Gemini 1.5 Pro improves on math tasks with more examples. 

\section{Sample Learning Ratio with Replacement}
\label{app:sample_learning_ratio_replacement}
\begin{figure*}[h]
    \centering
    \includegraphics[width=0.7\linewidth]{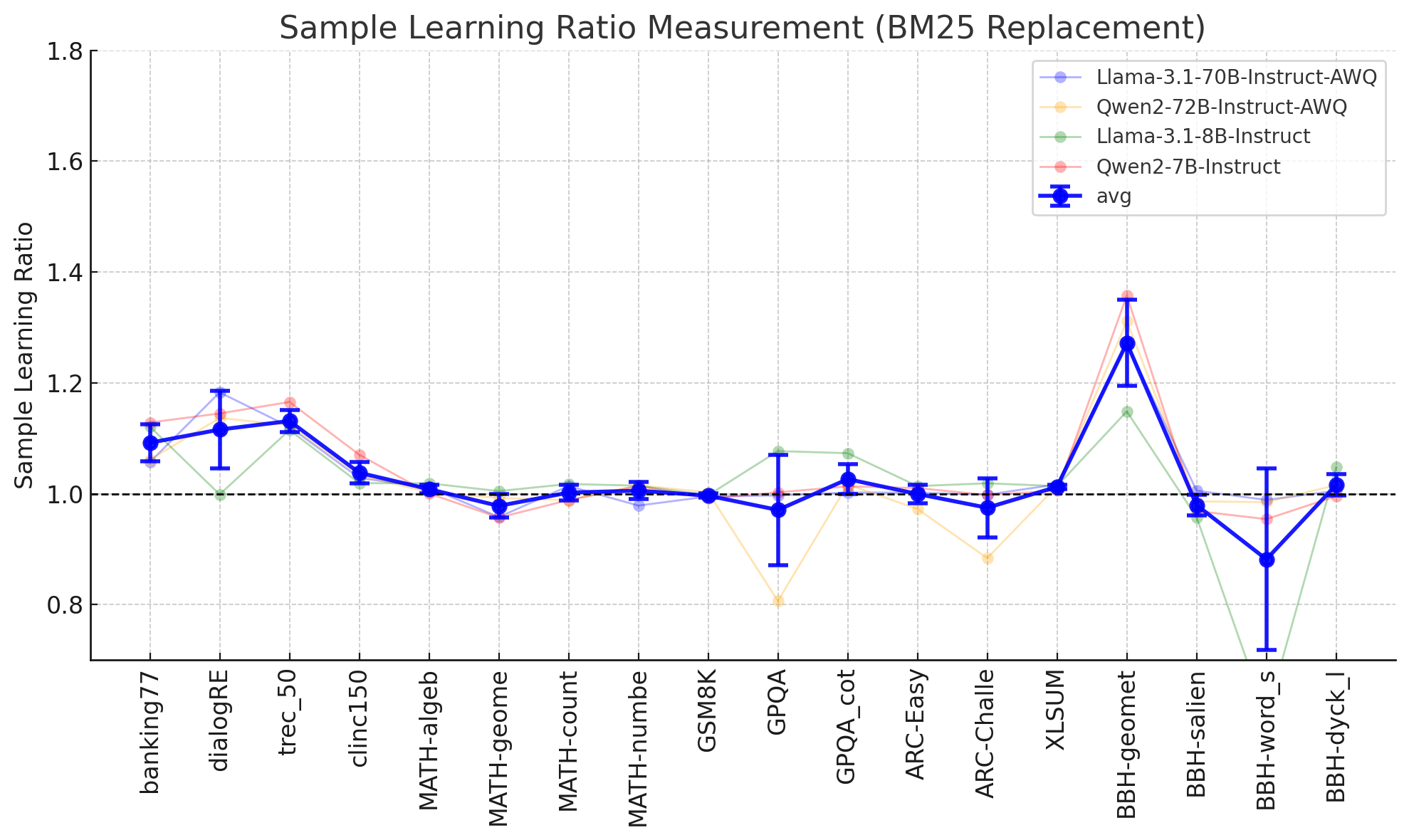}
    \caption{Sample Learning Ratio under the replacement setting with BM25 on all tasks expect XLSUM from 1k to 32k tokens. The ratio of 1 indicates models are not doing retrieval during ICL because similar demonstrations don't help models perform better. Similar to Figure \ref{fig:retrieval_load}, classification is the only category of tasks that has a higher ratio, which means classification tasks largely require model retrieval skills during ICL. The rest of the tasks are close to 1, and the models' performance on these tasks does not rely on retrieving similar examples.
    }
    \label{fig:retrieval_load_bm25_replace}
\end{figure*}

\begin{figure*}[h]
    \centering
    \includegraphics[width=0.7\linewidth]{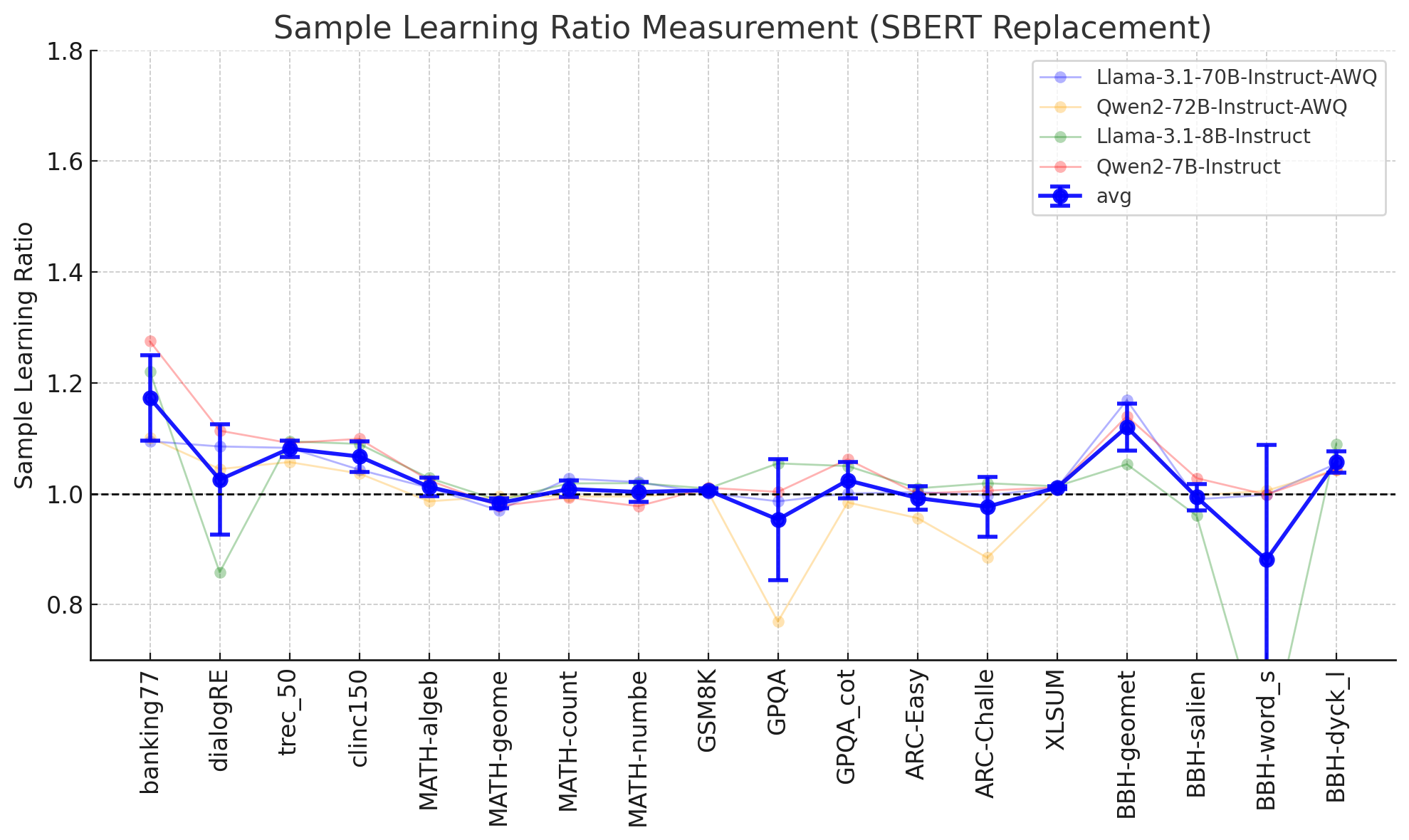}
    \caption{Sample Learning Ratio under the replacement setting with SBERT on selective tasks from 1k to 32k tokens. A ratio of 1 signifies that models do not perform retrieval during in-context learning (ICL), as similar demonstrations do not enhance their performance. As shown in Figure \ref{fig:retrieval_load}, classification tasks are the only category with a higher retrieval load ratio, indicating a strong dependence on retrieval during ICL. In contrast, other tasks exhibit ratios close to 1, suggesting minimal reliance on retrieval, with models' performance largely unaffected by retrieval-based demonstrations.
    }
    \label{fig:retrieval_load_sbert_replace}
\end{figure*}

To ensure the performance downgrade is not caused by the absence of certain labels in the experiment from Section 5, we replace similar examples with distant examples with the same labels. The new sample learning ratio formula is $\frac{score_{original}}{score_{replace}}$. We use Llama-3.1 and Qwen2 models and conduct this experiment from 1k to 32k with BM25 and from 1k to 32k with SBERT \citep{reimers-2019-sentence-bert} retrievers.

\textbf{BM25}: The trend in Figure \ref{fig:retrieval_load} matches the results of Figure \ref{fig:retrieval_load_bm25_replace}. All the classification tasks downgrade performance more when similar examples are replaced. However, the degree of downgrade is less significant than removing similar examples.

\textbf{SBERT}: For SentenceTransformer, we use all-MiniLM-L6-v2 as the base model. The trends observed from Figure \ref{fig:retrieval_load} and Figure \ref{fig:retrieval_load_bm25_replace} still hold in Figure \ref{fig:retrieval_load_sbert_replace}. That is, all the classification tasks still have a higher ratio and the non-classification tasks have a ratio close to 1.

\section{A deeper look into all-sample learning task}
\label{app:global context understanding}

\begin{figure*}[h]
    \centering
    \includegraphics[scale=0.4]{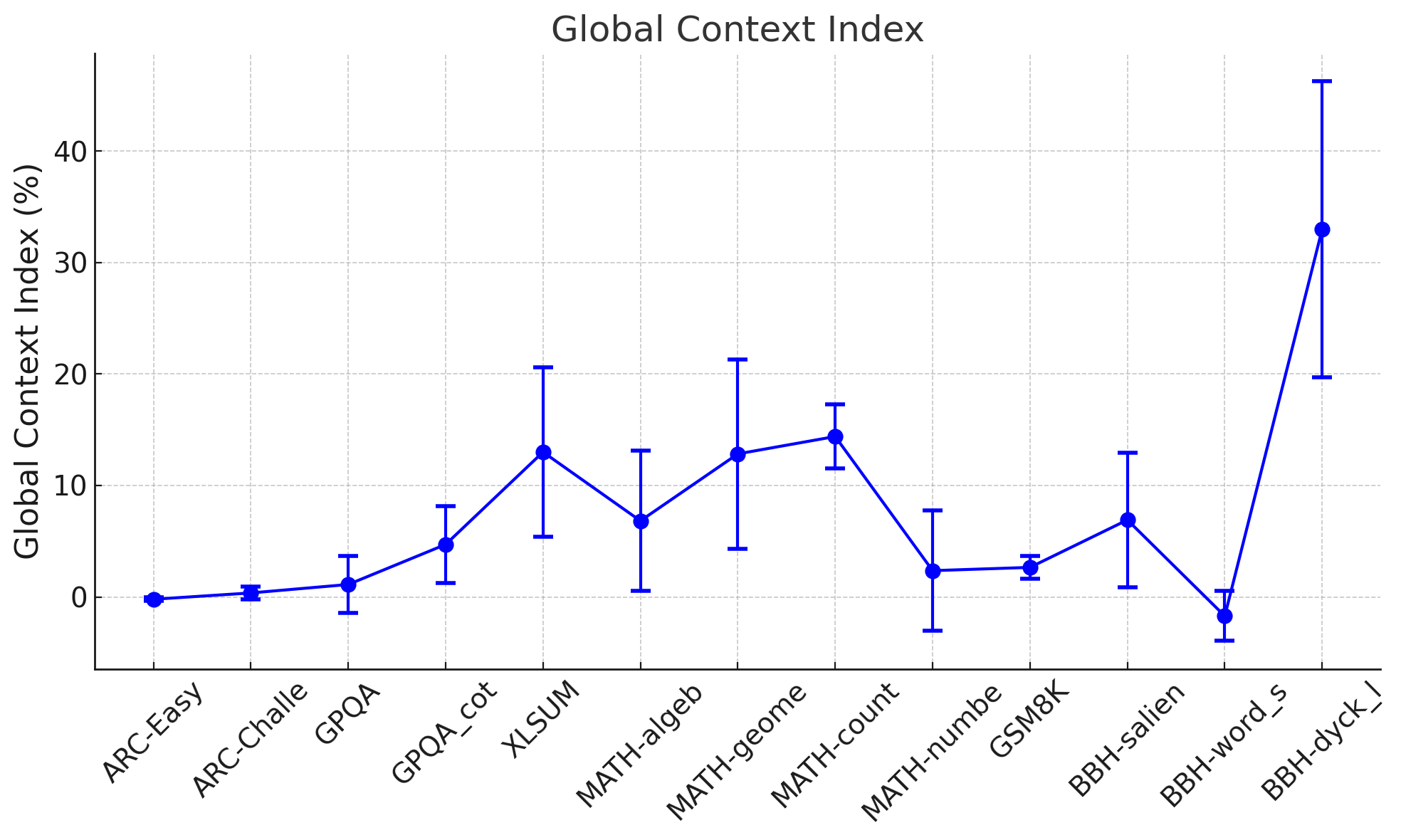}
    \caption{
    Global context index is the average \% difference between adding duplicated vs. unique examples from 2k to 64k context for non-retrieval tasks. 0\% means duplicating does not harm the model's performance. Easy tasks such as ARC and word sorting do not benefit from additional information. When a task is too difficult, e.g., GPQA, the model cannot effectively learn all demonstrations unless explanations are provided.
    }
    \label{fig:duplication}
\end{figure*}

In this section, we investigate which ASL tasks tend to benefit more from additional demonstrations and whether models use all the demonstrations to understand the task during ICL. 

To that end, we propose another metric, $\text{Global Context Index}$, to measure the global context understanding skill required by a task i.e., skills required to learn from all samples. Specifically, for each ASL task, we create two variants of demonstrations, both starting with the same demonstrations used in the 1k context-length experiment. For context lengths $l$ ranging from 2k to 64k, the \textit{unique variant} keeps adding unique demonstrations to the prompt, whereas the \textit{duplicate variant} repeats the same demonstrations as in the 1k context. At each length $l$, the performance of the unique variant is denoted as $\text{Perf}^{(l)}_{\text{unique}}$, and the performance of the duplicate variant as $\text{Perf}^{(l)}_{\text{duplicate}}$. We then average the percentage difference between $\text{Perf}^{(l)}_{\text{unique}}$ and $\text{Perf}^{(l)}_{\text{duplicate}}$ for $l = 2k$ to $l = 64k$ across 6 context lengths as:
\[
\text{Global Context Index} = \frac{1}{6}\sum_{l=2k}^{64k} \left(1 - \frac{\text{Perf}^{(l)}_{\text{duplicate}}}{\text{Perf}^{(l)}_{\text{unique}}}\right)
\]

If duplicating examples results in worse performance on an ASL task than adding unique examples, the global context index will be positive and suggests that the model benefits more from providing unique demonstrations. This means that performance improvements come from learning from diverse examples rather than simply picking up on formatting patterns or relying on spurious correlations between in-domain tokens and predictions. 
We use Llama-3.1-70B for the preliminary analysis because it is best at using additional demonstrations out of all models we have tested so far, e.g., it shows a high positive correlation between context lengths and performance in Figure \ref{fig:correlation}.

In Figure \ref{fig:duplication}, tasks such as the math problems, summarization, Dyck languages, translation error detection from BBH, and GPQA with explanations all have worse performance with duplicated demonstrations. This means that \textit{they necessitate a greater degree of global context understanding rather than relying on the retrieval of similar examples}. These tasks are often complex reasoning challenges, for which models may lack pretraining skills to solve perfectly, underscoring the need for additional demonstrations or deeper task comprehension.

ARC-Easy, ARC-Challenge, GPQA, and BBH word sorting tasks are indifferent to duplicating examples. This indicates that these tasks do not benefit from additional demonstrations. Most of these tasks assess the intrinsic abilities of the models reasoning with their parametric knowledge, thus a few demonstrations suffice.

Adding more demonstrations may introduce distractions rather than improve performance. Interestingly, GPQA with ``chain-of-thoughts'' benefit from additional examples. We suspect that without these solution steps, GPQA is too challenging for the model to understand even after seeing many demonstrations with answers only.

\textbf{9 Strong ASL Tasks.} Based on the analysis of the global context index, we remove ARC-Challenge and BBH word sorting tasks to create a new category of tasks called \textbf{strong ASL Tasks}. Evaluation results on strong ASL Tasks are summarized in Table \ref{tbl:strong_asl}.
\definecolor{improvement_high}{RGB}{255, 51, 51}   
\definecolor{improvement_medium}{RGB}{255, 102, 102} 
\definecolor{improvement_low}{RGB}{255, 204, 204}   

\definecolor{drop_high}{RGB}{51, 153, 255}         
\definecolor{drop_medium}{RGB}{102, 178, 255}      
\definecolor{drop_low}{RGB}{204, 229, 255}         

\begin{table*}[t]
    \centering

\resizebox{\linewidth}{!}{

    \begin{tabular}{lrrrrrrrrrr}
    \toprule
    \textbf{ASL Tasks} & \textbf{1k} & \textbf{2k} & \textbf{4k} & \textbf{8k} & \textbf{16k} & \textbf{32k} & \textbf{64k} & \textbf{128k} & \textbf{AVG.} & \textbf{AVG.L.} \\
    \midrule
    GLM-4-9b-Chat & 36.79 & \shade{36.79}{36.23} & \shade{36.79}{38.30} & \shade{36.79}{39.30} & \shade{36.79}{37.60} & \shade{36.79}{37.94} & \shade{36.79}{36.53} & \shade{36.79}{35.45} & 37.27 & 36.64 \\
    Mistral-Nemo-Instruct & 33.94 & \shade{33.94}{34.88} & \shade{33.94}{34.92} & \shade{33.94}{34.72} & \shade{33.94}{28.22} & \shade{33.94}{28.64} & \shade{33.94}{26.28} & \shade{33.94}{23.23} & 30.60 & 26.05 \\
    Mistral-Large-Instruct-AWQ & 57.09 & \shade{57.09}{56.30} & \shade{57.09}{56.21} & \shade{57.09}{56.12} & \shade{57.09}{56.43} & \shade{57.09}{53.33} & \shade{57.09}{42.98} & \shade{57.09}{13.10} & 48.94 & 36.47 \\
    Llama-3.1-8B-Instruct-AWQ & 31.31 & \shade{31.31}{32.79} & \shade{31.31}{33.02} & \shade{31.31}{34.50} & \shade{31.31}{34.25} & \shade{31.31}{35.22} & \shade{31.31}{33.71} & \shade{31.31}{27.88} & 32.84 & 32.27 \\
    Llama-3.1-70B-Instruct-AWQ & 45.53 & \shade{45.53}{47.60} & \shade{45.53}{48.39} & \shade{45.53}{49.08} & \shade{45.53}{49.64} & \shade{45.53}{49.83} & \shade{45.53}{47.74} & \shade{45.53}{13.88} & 43.99 & 37.23 \\
    Qwen2-7B-Instruct-AWQ & 37.75 & \shade{37.75}{39.47} & \shade{37.75}{43.86} & \shade{37.75}{44.55} & \shade{37.75}{42.83} & \shade{37.75}{35.17} & \shade{37.75}{33.00} & \shade{37.75}{32.70} & 38.67 & 33.62 \\
    Qwen2-72B-Instruct-AWQ & 47.38 & \shade{47.38}{49.03} & \shade{47.38}{50.32} & \shade{47.38}{50.69} & \shade{47.38}{50.78} & \shade{47.38}{48.56} & \shade{47.38}{48.18} & \shade{47.38}{48.68} & 49.20 & 48.47 \\
    Phi-3-Mini-Instruct & 29.86 & \shade{29.86}{29.20} & \shade{29.86}{26.61} & \shade{29.86}{26.95} & \shade{29.86}{27.65} & \shade{29.86}{26.34} & \shade{29.86}{25.54} & \shade{29.86}{23.08} & 26.90 & 24.98 \\
    Phi-3-Medium-Instruct & 37.74 & \shade{37.74}{37.15} & \shade{37.74}{31.49} & \shade{37.74}{32.02} & \shade{37.74}{33.04} & \shade{37.74}{33.19} & \shade{37.74}{33.06} & \shade{37.74}{24.56} & 32.78 & 30.27 \\
    Phi-3-Small-Instruct & 38.40 & \shade{38.40}{38.40} & \shade{38.40}{38.35} & \shade{38.40}{31.69} & \shade{38.40}{34.04} & \shade{38.40}{34.59} & \shade{38.40}{33.74} & \shade{38.40}{32.46} & 35.21 & 33.60 \\
    Jamba-1.5-Mini & 27.86 & \shade{27.86}{29.04} & \shade{27.86}{28.93} & \shade{27.86}{28.86} & \shade{27.86}{27.86} & \shade{27.86}{24.92} & \shade{27.86}{23.12} & \shade{27.86}{22.42} & 26.63 & 23.48 \\
    Gemini-1.5-Pro & 58.26 & \shadecell{58.26}{60.88} & \shadecell{58.26}{61.30} & \shadecell{58.26}{65.20} & \shadecell{58.26}{65.05} & \shadecell{58.26}{65.12} & \shadecell{58.26}{62.38} & \shadecell{58.26}{63.61} & \textbf{62.73} & \textbf{63.70} \\
\hline
    \bottomrule
    \end{tabular}
}
    \caption{
    Model performance on strong ASL tasks. AVG. is the average model performance of all context lengths. AVG.L. is the average model performance of 32k, 64k and 128k. \red{Red} indicates performance improvement compared to 1k.  \blue{
    Blue} indicates performance downgrade compared to 1k. A darker color means higher improvement or downgrade. \textbf{BOLD} number means the largest number of a column.}
    \label{tbl:strong_asl}
\end{table*}

\section{Task Performance}
\label{task performance}
\begin{figure*}[h]
    \centering
    \includegraphics[width=1\linewidth]{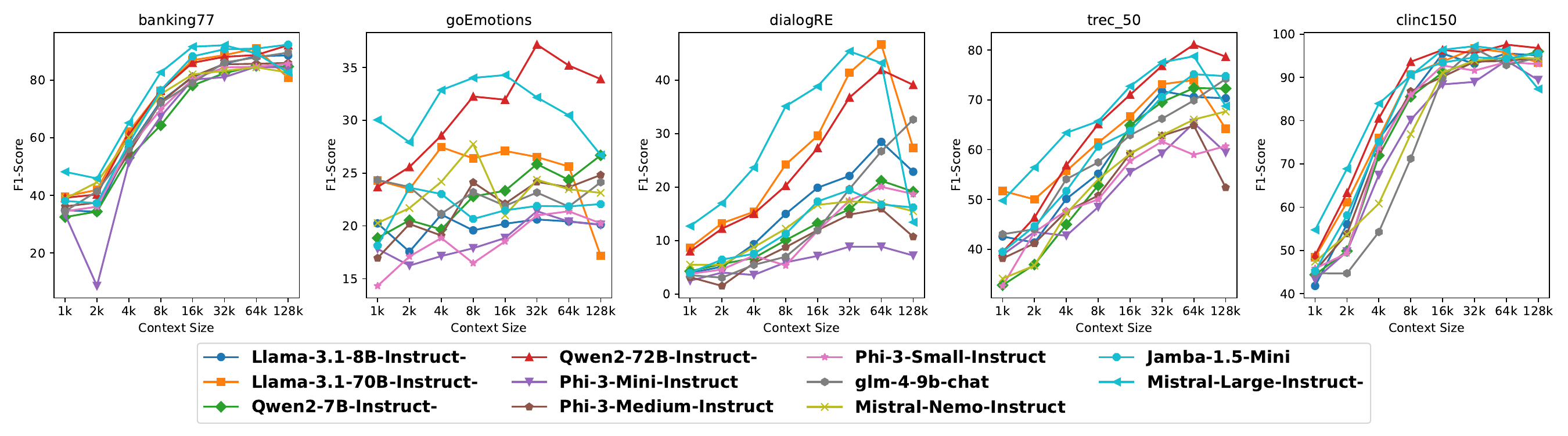}
    \caption{
    Models' performance on all classification tasks. All tasks except GoEmotions show a very consistent gain with increasing context size. We excluded GoEmotions from our benchmark because of the data's strong subjectivity. 
    }
    \label{fig:classification}
\end{figure*}

\begin{figure*}[h]
    \centering
    \includegraphics[width=1\linewidth]{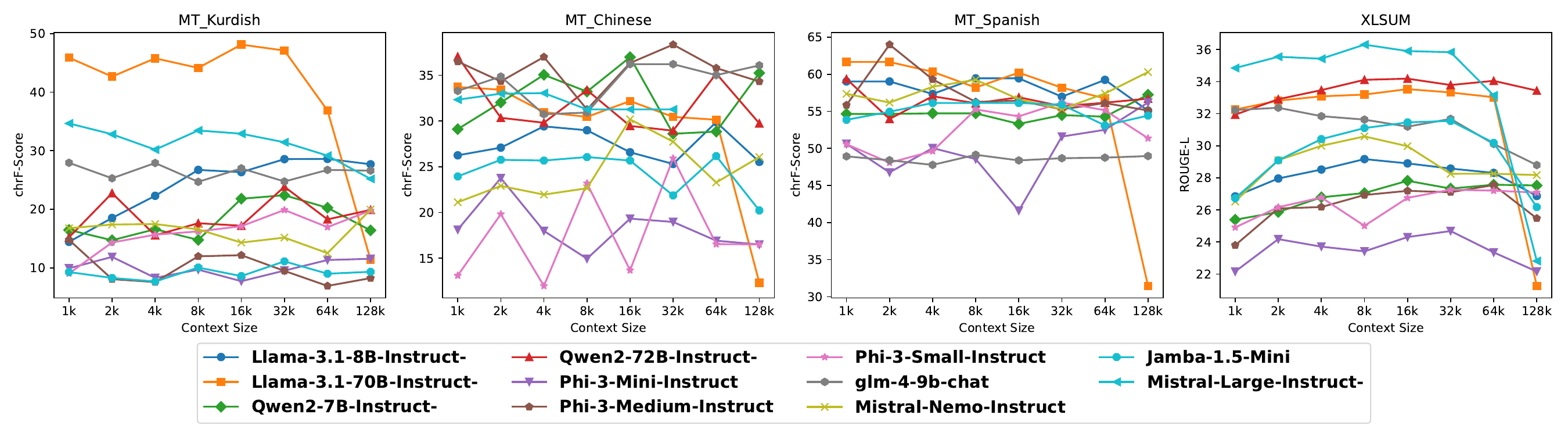}
    \caption{
    Models' performance on all translation tasks and the summarization task. For translation tasks, we do not observe a clear pattern among different languages and models, which can be caused by LCLMs' different multilingual abilities. We can see a slightly positive trend for the summarization task. 
    }
    \label{fig:translation}
\end{figure*}

\begin{figure*}[h]
    \centering
    \includegraphics[width=1\linewidth]{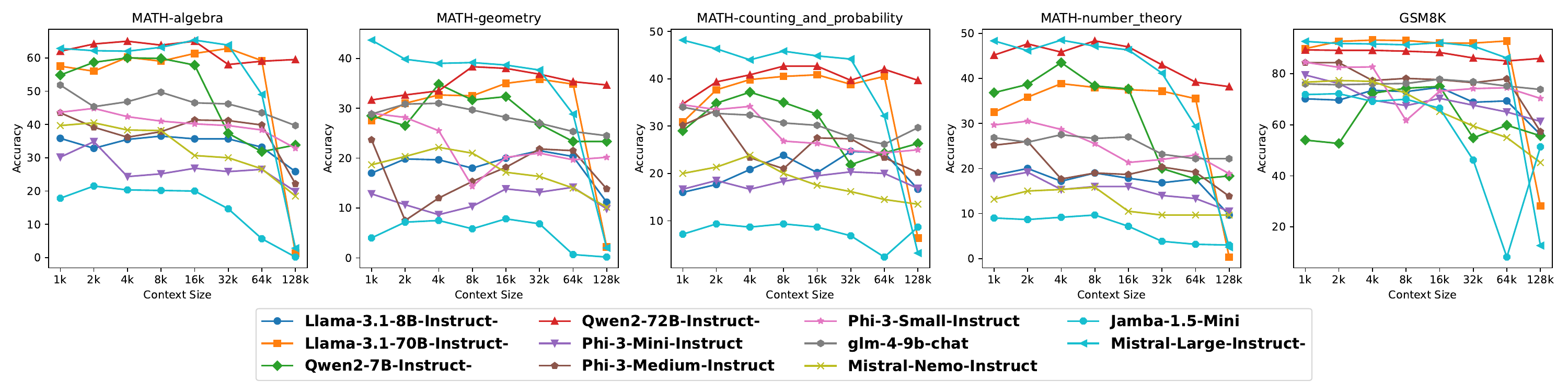}
    \caption{
    Models' performance on all math tasks. Overall, the larger and stronger models benefit more from the increasing context window size on math tasks. 
    }
    \label{fig:math}
\end{figure*}

\begin{figure*}[h]
    \centering
    \includegraphics[width=1\linewidth]{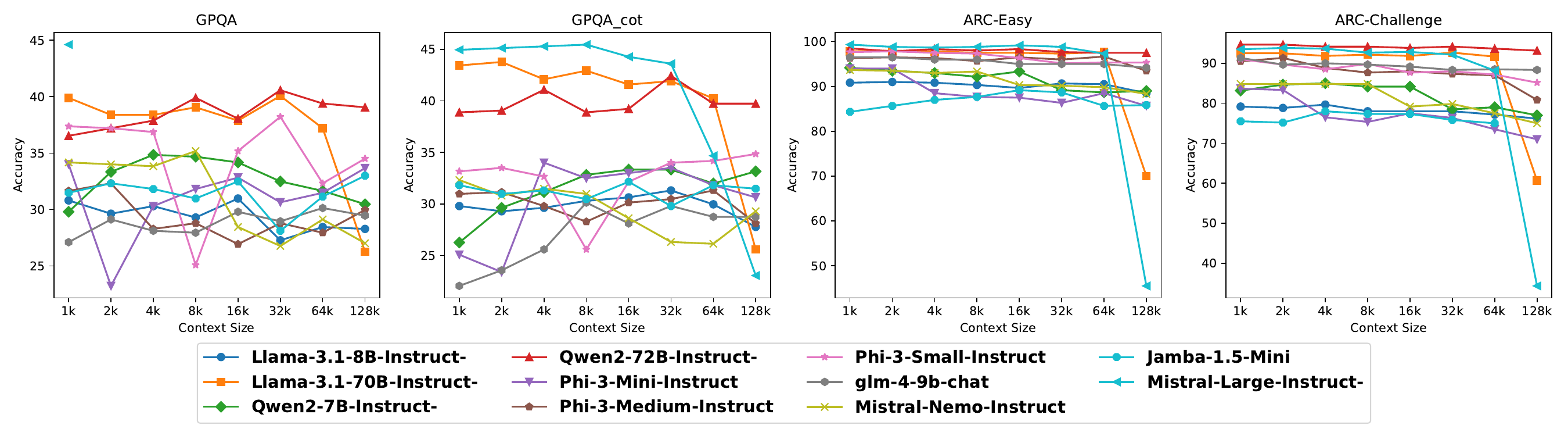}
    \caption{
    Models' performance on all science tasks. For the ARC task, the performance of all models stays the same across all context sizes. For GPQA, we can see larger and more robust LCLMs keep or increase their performance with the increasing context size. 
    }
    \label{fig:science}
\end{figure*}

\begin{figure*}[h]
    \centering
    \includegraphics[width=1\linewidth]{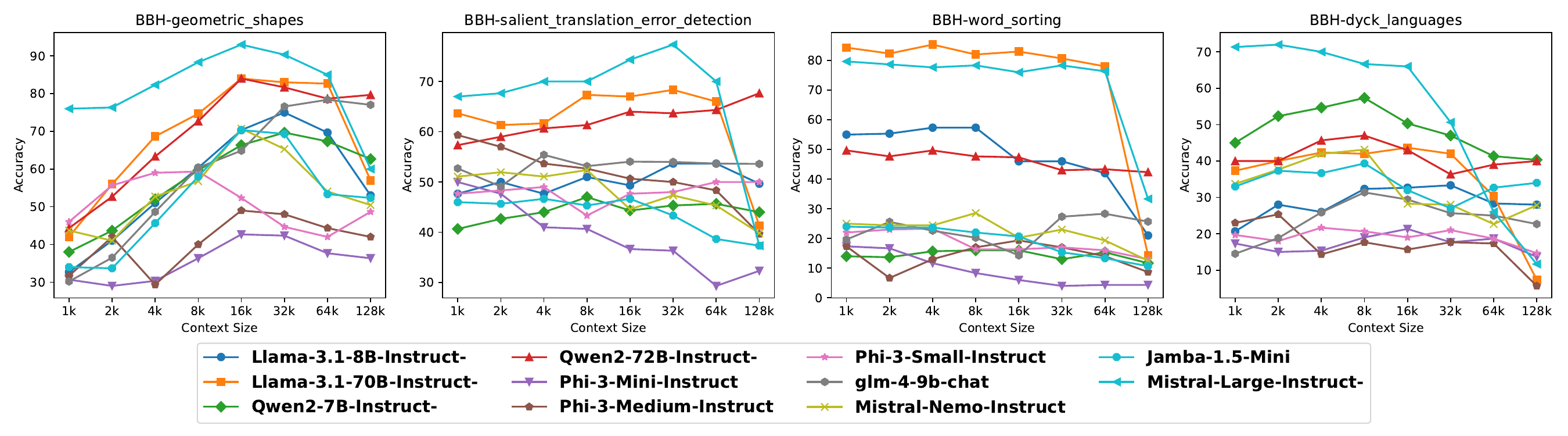}
    \caption{
    Models' performance on all symbolic tasks. For the geometric shape and translation error detection tasks, we can all model benefit from the increasing context length. We suspect the word sorting task may too easy for the models, so the lines are flat. For the dyck language task, the models experience performance gain up 16k context length but start downgrading afterward.  
    }
    \label{fig:symbolic}
\end{figure*}

In this section, we present the models' performance on individual tasks and group them by the task categories: classification (Figure \ref{fig:classification}), translation (Figure \ref{fig:translation}), summarization (Figure \ref{fig:classification}), and reasoning (Figures \ref{fig:math}, \ref{fig:science}, and \ref{fig:symbolic}).


\section{Error Analysis}
\label{app:error_analysis}
In this section, we include examples from MATH, BBH, and XLSUM tasks, highlighting the performance of Llama-3.1-70B and Qwen2-72B at both 16k and 128k context lengths. We analyze the strengths and shortcomings of each model’s outputs, focusing on accuracy, consistency, and the impact of extended context lengths on their reasoning processes. MATH (Table \ref{tbl:math_algebra}, \ref{tbl:math_counting}, and \ref{tbl:math_number_theory}), BBH-word\_sorting (Table \ref{tbl:bbh_word_1}, \ref{tbl:bbh_word_2}, and \ref{tbl:bbh_word_3}), BBH-dyck\_languages (Table \ref{tbl:bbh_dyck_1}, \ref{tbl:bbh_dyck_2}, and \ref{tbl:bbh_dyck_3}), and XLSUM (Table \ref{tbl:xlsum_example}).
\paragraph{Qwen2 and GLM-4 show relatively robust capabilities on both tasks.} 
The Qwen2-72B model consistently maintains performance across both SSL and ASL tasks, demonstrating its adaptability for longer contexts. Trained on data with up to 32k tokens, Qwen2 models employ modified RoPE frequency and training-free positional interpolation methods to handle longer contexts. However, the Qwen2 family models drop their performance from 16k to 32k in the ASL tasks but maintain their performance after 32k. This raises the question of whether the training-free length extension methods enable models to use additional demonstrations or merely maintain their performance in the short context length and ignore additional examples during many-shot ICL. 
Meanwhile, GLM-4-chat also shows a relatively robust performance at a longer context size and is the only model to experience a performance increase from 64k to 128k on SSL tasks. GLM-4's training methodology closely mirrors that of Llama 3.1 models, with adjustments to the RoPE base and continuous training on long-context data. The difference is, during SFT, GLM-4-9B follows LongAlign \citep{bai2024longalignrecipelongcontext}, which determines the length distribution of the long-context SFT data carefully. GLM-4-9B also goes through the RLHF stage with both short and long data.

\begin{figure*}[t!]
    \centering
    \begin{subfigure}[t]{0.5\linewidth}
        \centering
        \includegraphics[width=1\linewidth]{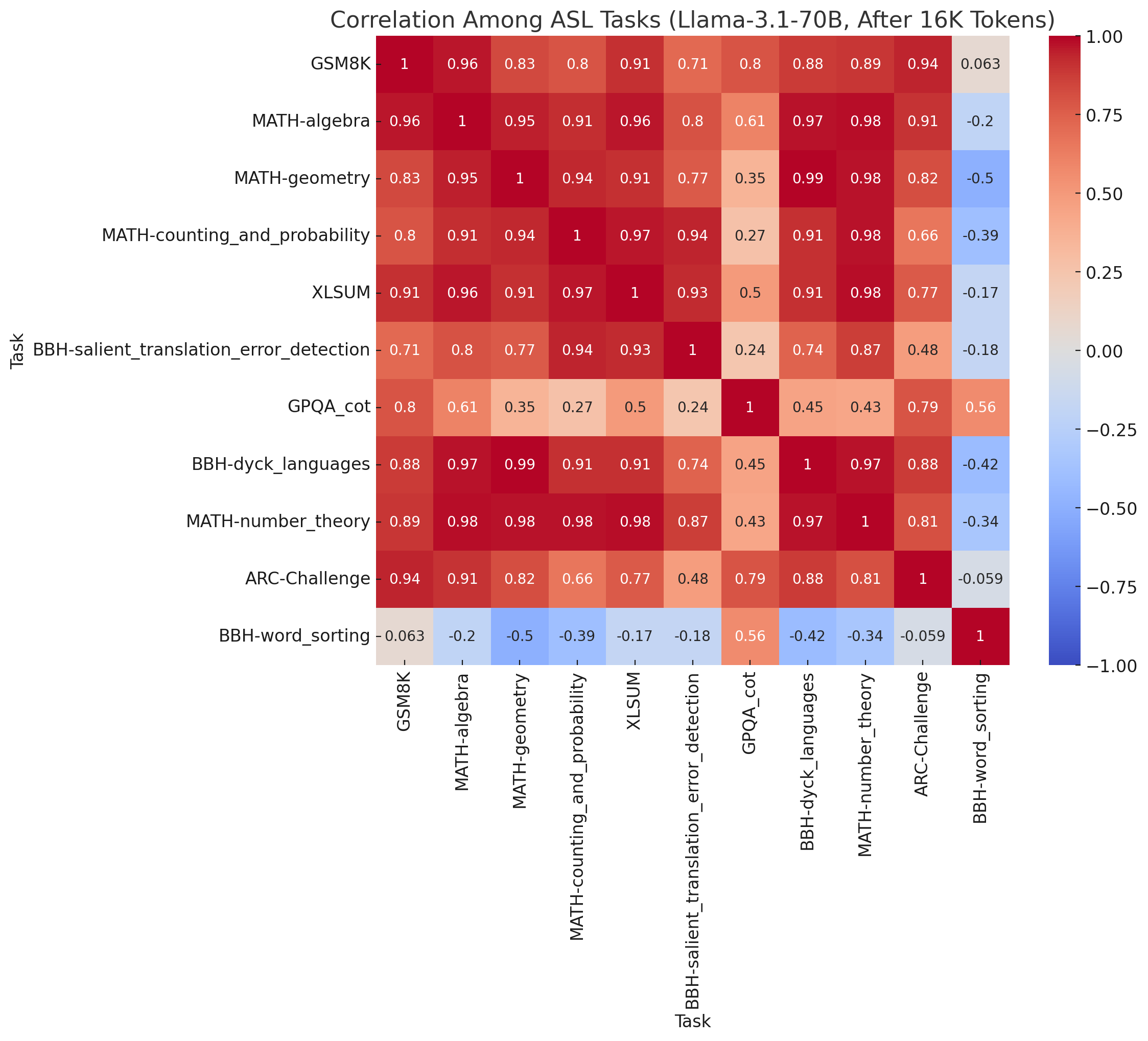}
        \caption{Llama-3.1-70B-Instruct's correlation}
    \end{subfigure}%
    ~ 
    \begin{subfigure}[t]{0.5\linewidth}
        \centering
        \includegraphics[width=1\linewidth]{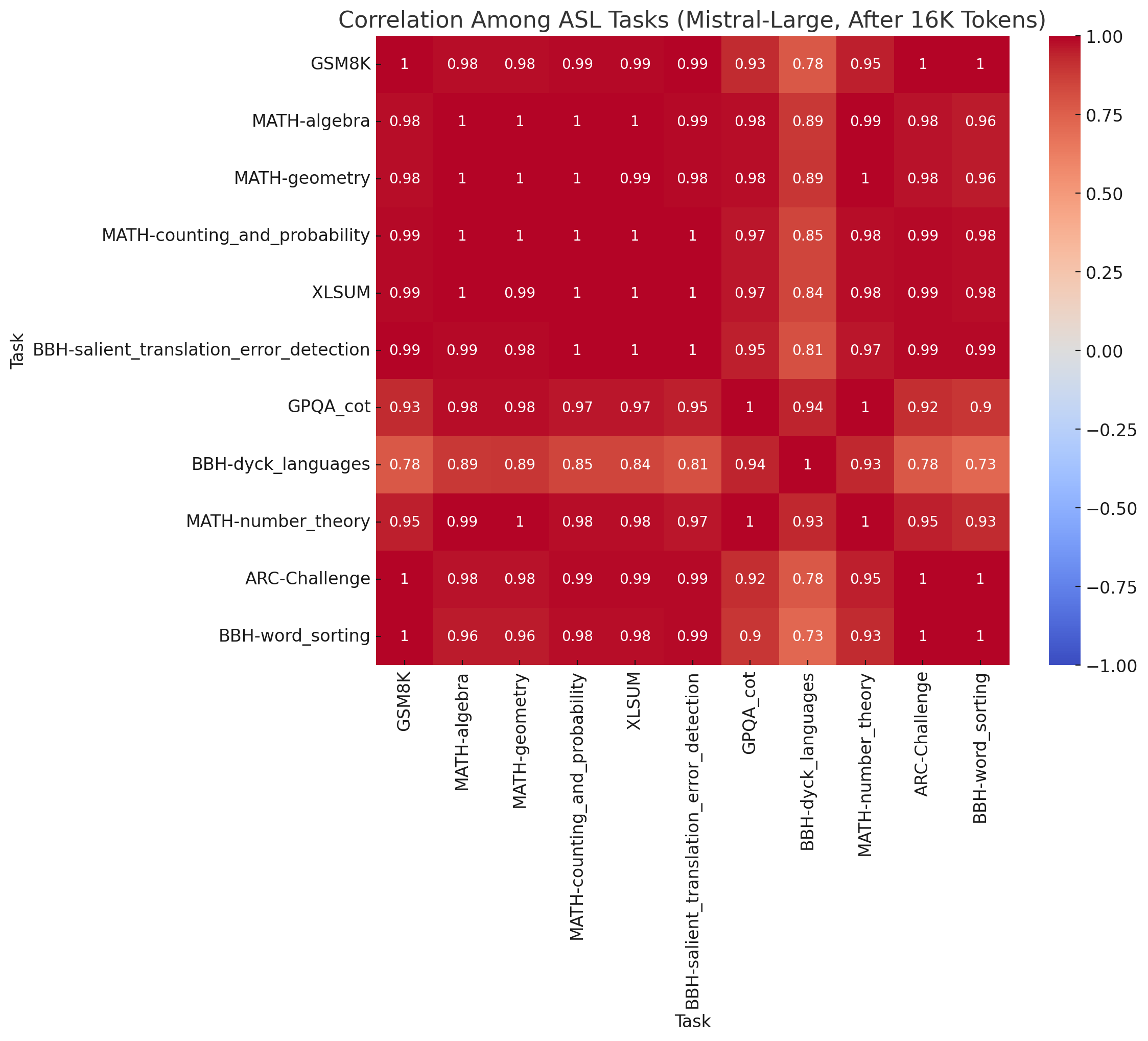}
        \caption{Mistral-Large's correlation}
    \end{subfigure}
    \begin{subfigure}[t]{0.5\linewidth}
        \centering
        \includegraphics[width=1\linewidth]{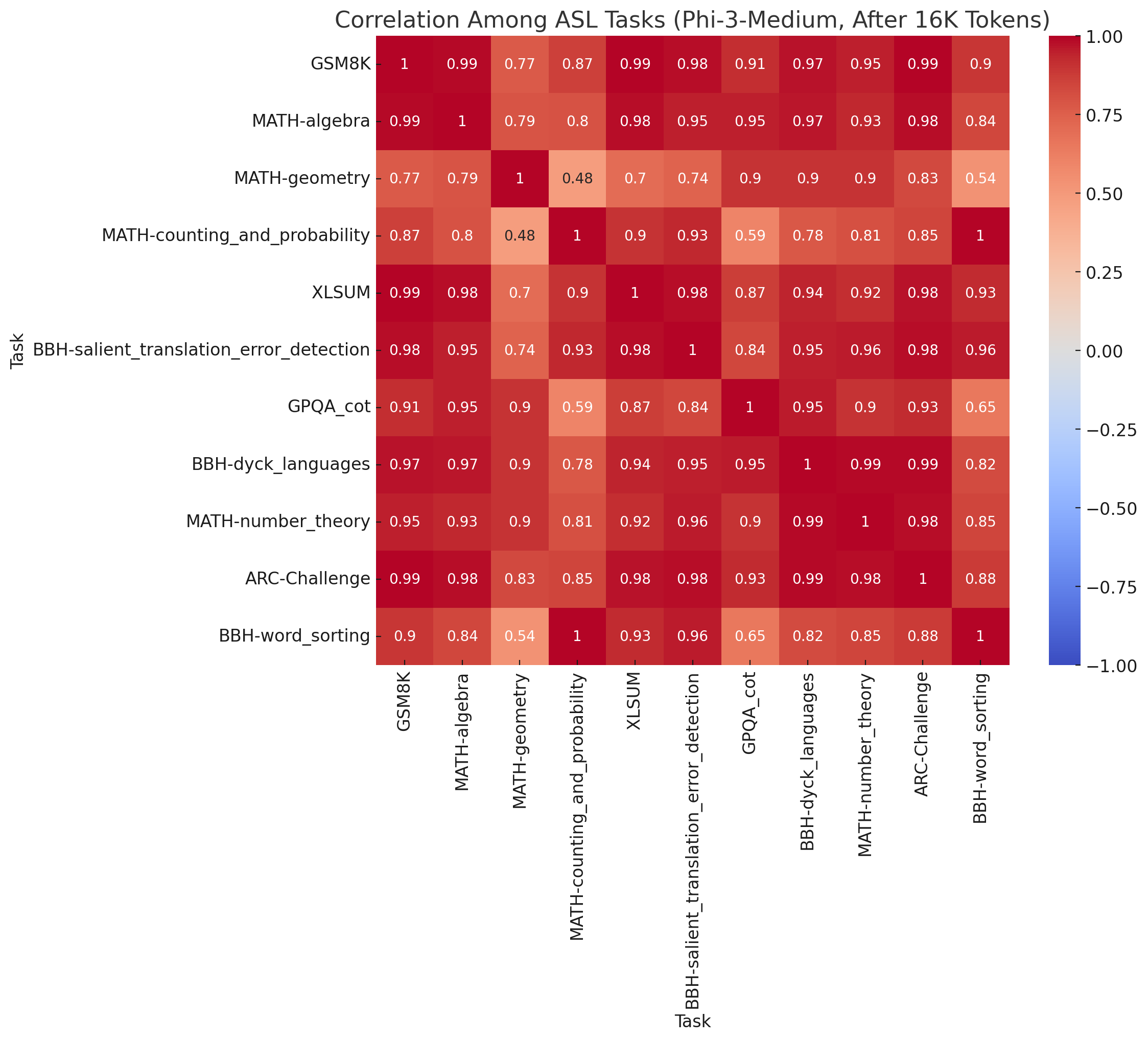}
        \caption{Phi-3-Medium's correlation}
    \end{subfigure}
    \caption{Correlation table among ASL tasks.}
    \label{fig:correlation_model}
\end{figure*}

\paragraph{Does LCLM's poor performance beyond 16k limit their potential in the world application?} In our ASL task evaluation, we distinguish between more practical tasks, such as XLSUM and MATH, and less real-world-relevant tasks, such as symbolic reasoning and ARC. To assess whether poor ASL performance beyond 16K tokens translates to real-world limitations, we analyzed the correlation between these tasks for three models that exhibit significant degradation after 16K tokens.
In Figure \ref{fig:correlation_model}, for Phi-3-Medium and Mistral-Large, we observed strong correlations across all tasks, which likely stems from their uniformly poor performance across the board. In contrast, Llama-3.1-70B exhibits a distinct pattern: while word sorting (a less practical task) does not correlate well with other tasks, all other ASL tasks show strong mutual correlations. Based on these findings, we believe that LCLMs' poor performance on ASL tasks beyond 16K tokens is a fair indicator of their limitations in real-world applications, where long-context understanding is often essential.
\begin{figure*}[t!]
    \centering
    \begin{subfigure}[t]{0.5\linewidth}
        \centering
        \includegraphics[width=1\linewidth]{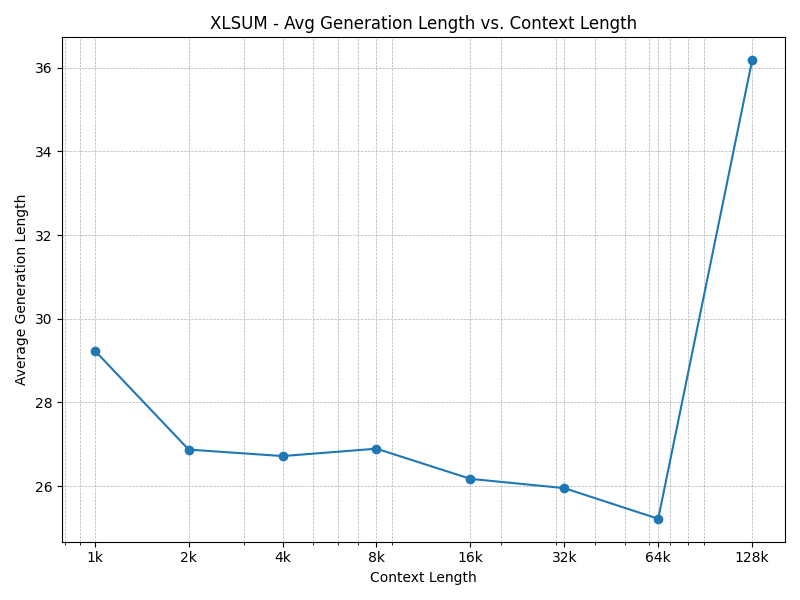}
        \caption{Llama-3.1-70B-Instruct's generation length on XLSUM}
    \end{subfigure}%
    ~ 
    \begin{subfigure}[t]{0.5\linewidth}
        \centering
        \includegraphics[width=1\linewidth]{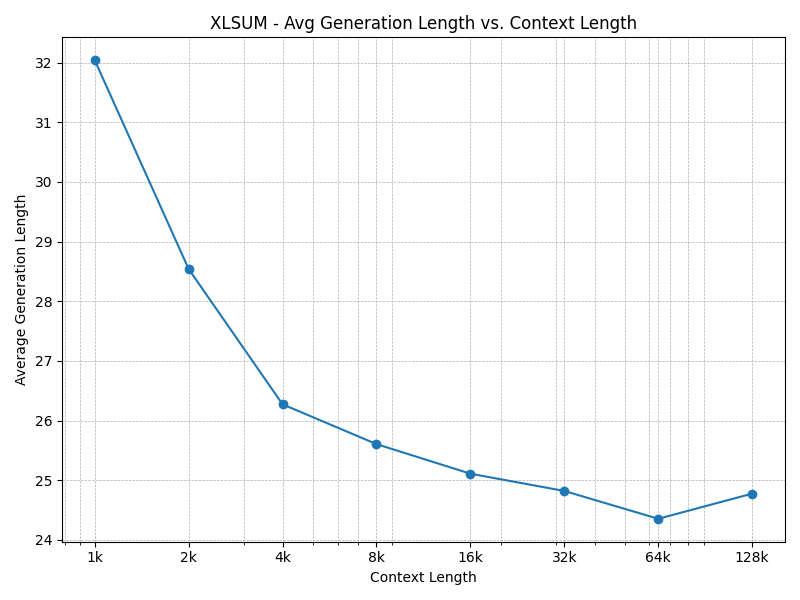}
        \caption{Qwen2-72B-Instruct's generation length on XLSUM}
    \end{subfigure}
    \caption{XLSUM generation length}
    \label{fig:XLSUM_length}
\end{figure*}
\begin{figure*}[t!]
    \centering
    \begin{subfigure}[t]{0.5\linewidth}
        \centering
        \includegraphics[width=1\linewidth]{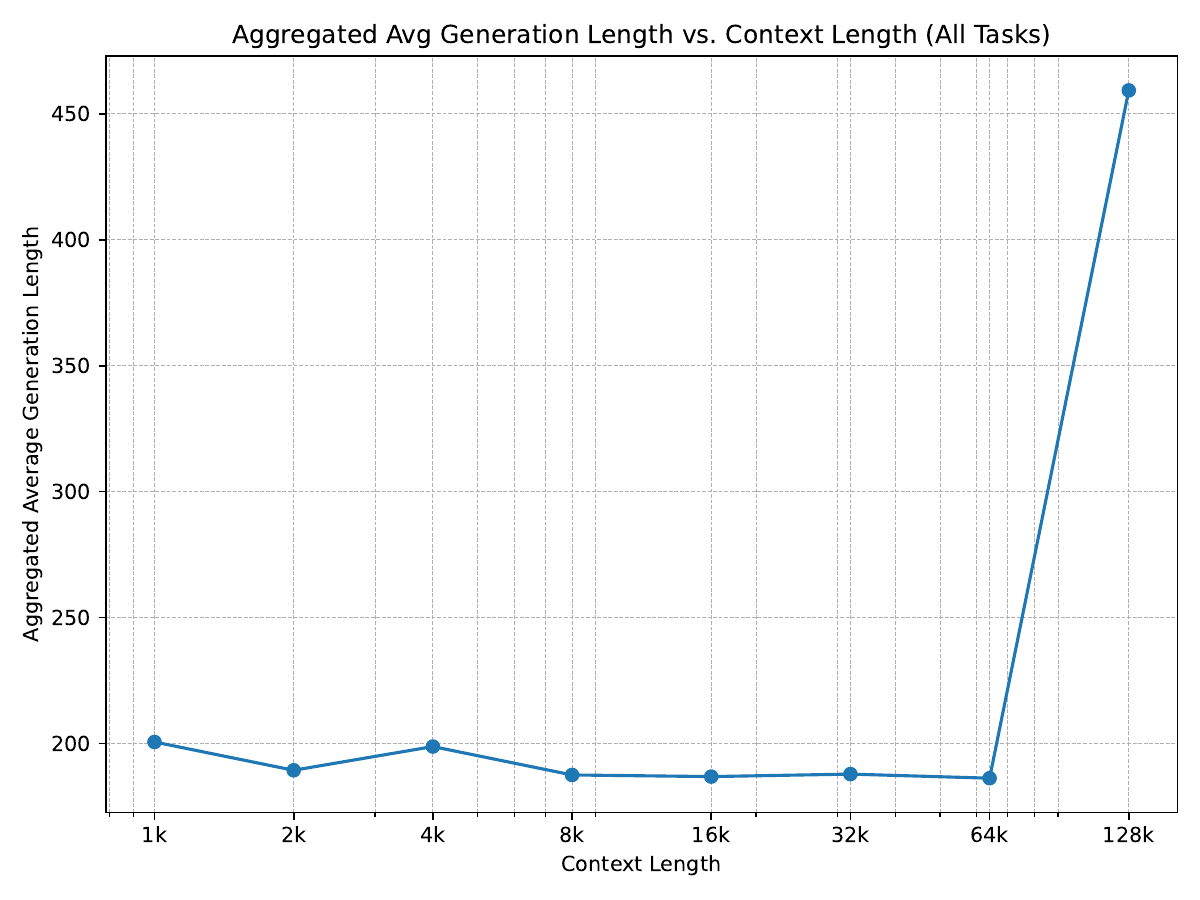}
        \caption{Llama-3.1-70B-Instruct's generation length on MATH}
    \end{subfigure}%
    ~ 
    \begin{subfigure}[t]{0.5\linewidth}
        \centering
        \includegraphics[width=1\linewidth]{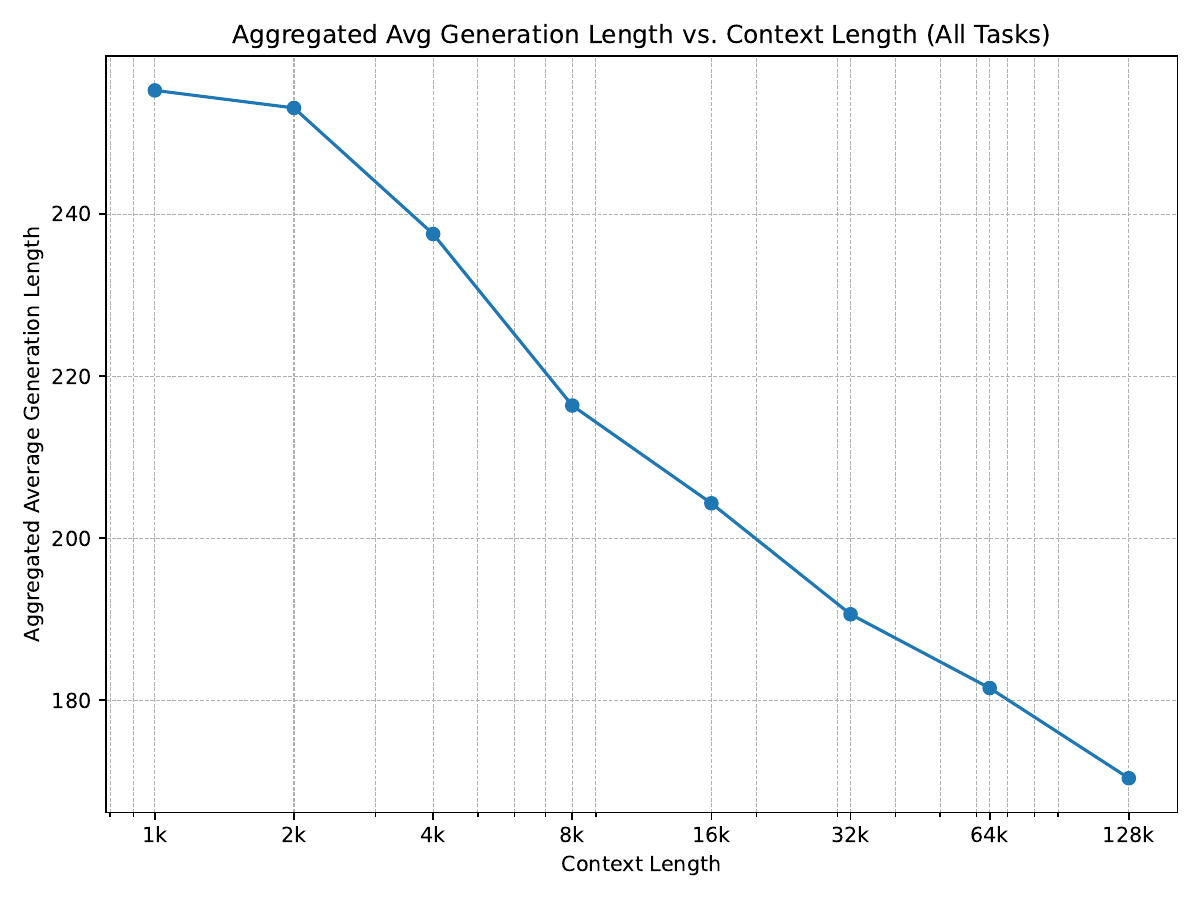}
        \caption{Qwen2-72B-Instruct's generation length on MATH}
    \end{subfigure}
    \caption{MATH generation length}
    \label{fig:MATH_length}
\end{figure*}
\clearpage

\onecolumn
\begin{longtable}{p{3cm} p{12cm}}
\toprule
\textbf{Task} & \textbf{Example} \\
\midrule
banking77 & Query: I have multiple of the same transaction\newline Intent: transaction\_charged\_twice \\
\midrule
goEmotions & Comment: It is bad tho.\newline Category: disapproval \\
\midrule
dialogRE & Dialogue: Speaker 1: I'm divorced!  I'm only 26 and I'm divorced! Speaker 2: Shut up! Speaker 3: You must stop! Speaker 1: That only took me an hour.\newline The list of 1 relations are (Speaker 1,26)\newline The respective relations between each entity pair are: per:age \\
\midrule
trec\_50 & Question: Describe the Finnish music personality Salonen 's appearance .\newline Type: DESC:desc \\
\midrule
clinc150 & Query: i need a good joke about office parties\newline Intent: tell\_joke \\
\midrule
MATH-algebra & Problem: How many cubic feet are in three cubic yards?\newline Solution: Cubing both sides of \(\displaystyle 1\text{ yard}=3\text{ feet}\) we find that 1 cubic yard equals 27 cubic feet. Therefore, 3 cubic yards are equal to \(27\cdot3=\boxed{81}\) cubic feet. \\
\midrule
MATH-geometry & Problem: In quadrilateral \(ABCD,\ BC=8,\ CD=12,\ AD=10,\) and \(m\angle A= m\angle B = 60^\circ.\) Given that \(AB = p + \sqrt{q},\) where \(p\) and \(q\) are positive integers, find \(p+q.\)\newline\newline Solution: [asy]draw((0,0)--(20.87,0)--(15.87,8.66)--(5,8.66)--cycle); draw((5,8.66)--(5,0)); draw((15.87,8.66)--(15.87,0)); draw((5,8.66)--(16.87,6.928)); label("\(A\)",(0,0),SW); label("\(B\)",(20.87,0),SE); label("\(E\)",(15.87,8.66),NE); label("\(D\)",(5,8.66),NW); label("\(P\)",(5,0),S); label("\(Q\)",(15.87,0),S); label("\(C\)",(16.87,7),E); label("12",(10.935,7.794),S); label("10",(2.5,4.5),W); label("10",(18.37,4.5),E); [/asy]\newline Draw line segment \(DE\) such that line \(DE\) is concurrent with line \(BC\). Then, \(ABED\) is an isosceles trapezoid so \(AD=BE=10\), and \(BC=8\) and \(EC=2\). We are given that \(DC=12\). Since \(\angle CED = 120^{\circ}\), using the Law of Cosines on \(\triangle CED\) gives 
\[
12^2=DE^2+4-2(2)(DE)(\cos 120^{\circ}),
\]
which simplifies to \(144-4=DE^2+2DE\). Adding 1 to both sides yields \((DE+1)^2=141\), so \(DE=\sqrt{141}-1\). In the \(30\!-\!60\!-\!90\) triangles \(\triangle DAP\) and \(\triangle EBQ\), we have \(AP=BQ=5\). Since \(PQ=DE\), it follows that 
\[
AB=AP+PQ+BQ=5+(\sqrt{141}-1)+5=9+\sqrt{141},
\]
so \((p,q)=(9,141)\) and \(p+q=\boxed{150}\). \\
\midrule
MATH-counting & Problem: Sarah wants to order a pizza with 2 different toppings. She has 9 different toppings to choose from. How many different pizzas could she order?\newline Solution: There are 9 choices of toppings, and we need to choose 2 distinct toppings. This is represented by the number of 2-element subsets of a 9-element set. We use the binomial coefficient ${9 \choose 2} = \boxed{36}$ to compute this. \\
\midrule
MATH-number\_theory & Problem: Express eleven in base 2.\newline Solution: We have \(11=1\cdot 2^3 + 0\cdot 2^2 + 1\cdot 2^1 + 1\cdot 2^0,\) so \(11=\boxed{1011\_2}\). \\
\midrule
BBH-geometric\_shapes & Input: This SVG path element \texttt{<path d="M 53.64,29.71 L 61.55,33.55 M 61.55,33.55 L 65.49,42.75 M 65.49,42.75 L 60.75,49.85 M 60.75,49.85 L 54.92,52.75 L 48.80,50.52 M 48.80,50.52 L 44.97,43.03 M 44.97,43.03 L 46.15,33.55 M 46.15,33.55 L 53.64,29.71"/>} draws a\newline Options: (A) circle, (B) heptagon, (C) hexagon, (D) kite, (E) line, (F) octagon, (G) pentagon, (H) rectangle, (I) sector, (J) triangle\newline Target: (F) \\
\midrule
BBH-salient\_translation\_ \\
error\_detection & Input: The following translations from German to English contain a particular error. That error will be one of the following types: Named Entities (an entity is changed), Numerical Values (values or units are changed), Modifiers or Adjectives (modifiers are changed), Negation or Antonyms (negations or opposites are altered), Facts (trivial factual errors), Dropped Content (significant content is removed). Please identify that error.  \newline Source: Die unvollständige Liste der Baudenkmale in Barsinghausen enthält Baudenkmale der Barsinghausener Kernstadt sowie der Ortsteile Bantorf, Barrigsen, Eckerde, Egestorf, Göxe, Großgoltern, Groß Munzel, Hohenbostel, Holtensen, Kirchdorf, Landringhausen, Langreder, Nordgoltern, Ostermunzel, Stemmen, Wichtringhausen und Winninghausen.\newline Translation: The complete list of architectural monuments in Barsinghausen contains architectural monuments of the Barsinghausen core town as well as the districts bantorf, Barrigsen, Eckerde, Egestorf, Göxe, Großgoltern, Groß Munzel, Hohenbostel, Holtensen, Kirchdorf, Landringhausen, Langreder, Nordgoltern, Ostermunzel, Stemmen, Wichtringhausen and Winninghausen.\newline The translation contains an error pertaining to\newline Options: (A) Modifiers or Adjectives, (B) Numerical Values, (C) Negation or Antonyms, (D) Named Entities, (E) Dropped Content, (F) Facts\newline Target: (C) \\
\midrule
BBH-word\_sorting & Input: Sort the following words alphabetically: List: thrill splutter panicking scorch same dot prod obstetric malton onus drumhead delmarva barn embezzle it\&t damp guru subsist entirety greene\newline Target: barn damp delmarva dot drumhead embezzle entirety greene guru it\&t malton obstetric onus panicking prod same scorch splutter subsist thrill \\
\midrule
BBH-dyck\_languages & Input: Complete the rest of the sequence, making sure that the parentheses are closed properly. Input: ( [ [ [ { } ] ] { < [ < [ { } ] > ] > }\newline Target: ] ) \\
\midrule
GPQA & Question: Determine which set of states mentioned below are only entangled states:\newline (a) \(\frac{1}{\sqrt{30}}\)(\(|00\rangle+ 2i|01\rangle- 3|10\rangle- 4i|11\rangle\))\newline (b) \(\frac{1}{5}\)(\(|00\rangle+ 2i|01\rangle- 2|10\rangle- 4i|11\rangle\))\newline (c) \(\frac{1}{2}\)(\(|00\rangle+ |01\rangle+|10\rangle- |11\rangle\))\newline (d) \(\frac{1}{2}\)(\(|00\rangle+ |01\rangle-|10\rangle- |11\rangle\))\newline Options: A. a,b \quad B. b,d \quad C. c,d \quad D. a,c\newline Answer: D \\
\midrule
GPQA\_cot & Question: Determine which set of states mentioned below are only entangled states:\newline (a) \(\frac{1}{\sqrt{30}}\)(\(|00\rangle+ 2i|01\rangle- 3|10\rangle- 4i|11\rangle\))\newline (b) \(\frac{1}{5}\)(\(|00\rangle+ 2i|01\rangle- 2|10\rangle- 4i|11\rangle\))\newline (c) \(\frac{1}{2}\)(\(|00\rangle+ |01\rangle+|10\rangle- |11\rangle\))\newline (d) \(\frac{1}{2}\)(\(|00\rangle+ |01\rangle-|10\rangle- |11\rangle\))\newline Options: A. a,b \quad B. b,d \quad C. c,d \quad D. a,c\newline Answer: D\newline Explanation: For a state \(a|00\rangle+ b|01\rangle+ c|10\rangle+ d|11\rangle\), separability requires \(a\cdot d = b\cdot c\). This condition is not met for option a,c; hence both states are entangled. \\
\midrule
ARC-Challenge & Question: One important difference between living things and nonliving things is that only living things have\newline 1. compounds \quad 2. elements \quad 3. molecules \quad 4. cells\newline Answer: 4 \\
\midrule
ARC-Easy & Question: Infections may be caused by\newline 1. mutations \quad 2. microorganisms \quad 3. toxic substances \quad 4. climate changes\newline Answer: 2 \\
\midrule
MT\_Kurdish & English: Every corner store is filled with a confusing array of pre-paid phone cards that can be used from pay phones or ordinary telephones.\newline\newline Kurdish: ... \\
\midrule
MT\_Chinese & English: Every corner store is filled with a confusing array of pre-paid phone cards that can be used from pay phones or ordinary telephones.\newline\newline Chinese: ... \\
\midrule
MT\_Spanish & English: Every corner store is filled with a confusing array of pre-paid phone cards that can be used from pay phones or ordinary telephones.\newline\newline Spanish: En las tiendas de cada esquina hay una enorme y confusa variedad de tarjetas de telefonía prepagas que pueden utilizarse en teléfonos públicos o comunes. \\
\midrule
GSM8K & Question: If one Burmese python can eat one 50-cm alligator per week, how many Burmese pythons would it take to eat fifteen 50-centimeter alligators in three weeks?\newline Answer: 15 alligators eaten over 3 weeks gives a rate of \(15/3=5\) alligators per week. Thus, at 1 alligator per week per python, it would take 5 pythons to eat 15 alligators.\newline \(\boxed{5}\) \\
\bottomrule
\caption{Input and output example from each task}
\label{tbl:example_prompts}
\end{longtable}
\twocolumn

\onecolumn
\begin{longtable}{p{15cm}}
\small \\
\toprule 
\textbf{Task:} Math-Algebra \\
\midrule
\textbf{Problem}: \\
A triangle has vertices at coordinates $(11,1)$, $(2,3)$ and $(3,7)$. What is the number of units in the length of the longest side of the triangle? \\
\midrule
\textbf{Difficulty}: Level 3 \\
\midrule
\textbf{Solution}: \\
We must find the distance between each pair of points by using the distance formula.
The distance between $(11, 1)$ and $(2, 3)$ is $\sqrt{(11-2)^2+(1-3)^2}=\sqrt{81+4}=\sqrt{85}$.
The distance between $(2, 3)$ and $(3, 7)$ is $\sqrt{(2-3)^2+(3-7)^2}=\sqrt{1+16}=\sqrt{17}$.
The distance between $(3, 7)$ and $(11, 1)$ is $\sqrt{(11-3)^2+(1-7)^2}=\sqrt{64+36}=\sqrt{100}=10$.
$10$ is larger than $\sqrt{85}$ and $\sqrt{17}$. Thus, the longest side of the triangle has length $\boxed{10}$. \\
\midrule
\textbf{Model}: Llama-3.1-70B \\
\midrule
\textbf{16k Solution}: \\
The longest side of the triangle is the side with the greatest length. The length of a side can be found using the distance formula, which is $$\sqrt{(x_2-x_1)^2+(y_2-y_1)^2}.$$  
We can use this formula to find the lengths of the three sides of the triangle. The first side has a length of $$\sqrt{(11-2)^2+(1-3)^2}=\sqrt{9^2+2^2}=\sqrt{85}.$$  
The second side has a length of $$\sqrt{(11-3)^2+(1-7)^2}=\sqrt{8^2+6^2}=\sqrt{100}=10.$$  
The third side has a length of $$\sqrt{(2-3)^2+(3-7)^2}=\sqrt{1^2+4^2}=\sqrt{17}.$$  
The longest side of the triangle is the side with the greatest length, which is the second side with a length of $\boxed{10}$ units. \\
\midrule
\textbf{128k Solution}: \\
We use the distance formula:
\[
\begin{aligned}
    \sqrt{(-1-2)^2+(2-3)^2+(3-7)^2} &= \\
    \vdots \textcolor{red}{\ Repeat\ 13\ times} \\
    \sqrt{(-1-2)^2+(2-3)^2+(3-7)^2} &= \boxed{10}.
\end{aligned}
\]
\\
\midrule
\textbf{Model}: Qwen2-72B \\
\midrule
\textbf{16k Solution}: \\
We must find the distance between each pair of points.\\[0.5em]
The distance between $(11,1)$ and $(2,3)$ is $\sqrt{(11-2)^2+(1-3)^2}=\sqrt{81+4}=\sqrt{85}$.\\[0.5em]
The distance between $(11,1)$ and $(3,7)$ is $\sqrt{(11-3)^2+(1-7)^2}=\sqrt{64+36}=\sqrt{100}=10$.\\[0.5em]
The distance between $(2,3)$ and $(3,7)$ is $\sqrt{(2-3)^2+(3-7)^2}=\sqrt{1+16}=\sqrt{17}$.\\[0.5em]
Thus, the longest side of the triangle has length $\boxed{10}$.\\
\midrule
\textbf{128k Solution}:\\
We find the distance between each pair of points. The distance between $(11,1)$ and $(2,3)$ is $\sqrt{(11-2)^2+(1-3)^2}=\sqrt{81+4}=\sqrt{85}$. The distance between $(11,1)$ and $(3,7)$ is $\sqrt{(11-3)^2+(1-7)^2}=\sqrt{64+36}=\sqrt{100}=10$. The distance between $(2,3)$ and $(3,7)$ is $\sqrt{(2-3)^2+(3-7)^2}=\sqrt{1+16}=\sqrt{17}$. Therefore, the longest side of the triangle has length $\boxed{10}$.\\
\midrule
\textbf{Analysis}: At 128k, Llama-3.1-70B misapplies the distance formula and repeats the same incorrect computation multiple times. Although the final boxed answer is correct, the reasoning is completely flawed. In contrast, Qwen2-72B maintains consistent reasoning in both 16k and 128k solutions and arrives at the correct answer $\boxed{10}$. \\
\bottomrule
\caption{An example problem from MATH\_algebra}
\label{tbl:math_algebra}
\end{longtable}
\twocolumn
\onecolumn
\begin{longtable}{p{15cm}}
\small \\
\toprule
\textbf{Task:} MATH-counting\_and\_probability \\
\midrule
\textbf{Problem:} \\
Sarah wants to order a pizza with 2 different toppings. She has 9 different toppings to choose from. How many different pizzas could she order? \\
\midrule
\textbf{Difficulty:} Level 1 \\
\midrule
\textbf{Solution:} \\
There are 9 choices of toppings, and we need to choose 2 distinct toppings. This is represented by the number of 2-element subsets of a 9-element set. We use the binomial coefficient ${9 \choose 2} = \boxed{36}$ to compute this. \\
\midrule
\textbf{Model:} Llama-3.1-70B \\
\midrule
\textbf{16k Solution:} \\
There are $\binom{9}{2} = \boxed{36}$ different pizzas she could order. \\
\midrule
\textbf{Solution at 128k:} \\
Sarah can choose 2 different toppings from 9 different toppings. There are $\binom{9}{2} = 42$ ways to choose 2 different toppings. For each of these 42 different toppings, there are 2 different pizzas, one with the first topping first and the other with the second topping first. Therefore, there are $42 \cdot 2 = \boxed{84}$ different pizzas. \\
\midrule
\textbf{Model:} Qwen2-72B \\
\midrule
\textbf{16k Solution:} \\
Sarah can choose 2 toppings from 9 in $\binom{9}{2} = \boxed{36}$ ways. \\
\midrule
\textbf{Solution at 128k:} \\
Sarah can choose her first topping in 9 ways. For the second topping, she has 8 choices left. So, there are $9\times 8=72$ ways to choose the toppings. However, since the order of the toppings does not matter, we have counted each pair of toppings twice. Therefore, the number of different pizzas she can order is $\dfrac{9\times 8}{2}=\boxed{36}$. \\
\midrule
\textbf{Analysis:} \\
Llama-3.1-70B's 128k solution incorrectly computes the number of ways to choose 2 toppings by mistakenly taking $\binom{9}{2}$ as 42 and then multiplying by 2, resulting in an incorrect answer of $\boxed{84}$. In contrast, Qwen2-72B's 128k solution correctly accounts for the unordered nature of the topping selection, arriving at the correct answer of $\boxed{36}$. \\
\bottomrule
\caption{An example problem from MATH\_counting\_and\_probability}
\label{tbl:math_counting}
\end{longtable}
\twocolumn

\onecolumn
\begin{longtable}{p{15cm}}
\small \\
\toprule
\textbf{Task:} MATH-number\_theory \\
\midrule
\textbf{Problem:} \\
Tim is doing a cakewalk with $11$ steps. He takes his first step on step $1$ and takes a total of $139$ steps, walking in a circle (so that after the 11th step he reaches the first step again). Which step of the cakewalk does he end on? \\
\midrule
\textbf{Difficulty:} Level 2 \\
\midrule
\textbf{Solution:} \\
When you divide $139$ steps by $11$ steps, you get $12$ with a remainder of $7$. Therefore, Tim lands on step $\boxed{7}$. \\
\midrule
\textbf{Model:} Llama-3.1-70B \\
\midrule
\textbf{16k Solution:} \\
We can write $139 = 11 \cdot 12 + 7$. Thus, after $12$ full cycles, he will end up $7$ steps from the first step. Since $7$ is less than $11$, he will end up on the $7^\text{th}$ step of the cakewalk. Thus, the answer is $\boxed{7}$. \\
\midrule
\textbf{Solution at 128k:} \\
Since $139$ is divisible by $11$, the number of steps is a multiple of $11$. Since $139$ is less than $200$, the number of steps is less than $200$ and therefore is $139$ or $139-1$ or $139-2$ or $139-3$ or $139-4$ or $139-5$ or $139-6$ or $139-7$ or $139-8$ or $139-9$ or $139-10$ or $139-11$ or $139-12$ or $139-13$ or $139-14$ or $139-15$ or $139-16$ or $139-17$ or $139-18$ or $139-19$ or $139-20$ or $139-21$ or $139-22$ or $139-23$ or $139-24$ or $139-25$ or $139-26$ or $139-27$ or $139-28$ or $139-29$ or $139-30$ or $139-31$ or $139-32$ or $139-33$ or $139-34$ or $139-35$ or $139-36$ or $139-37$ or $139-38$ or $139-39$ or $139-40$ or $139-41$ or $139-42$ or $139-43$ or $139-44$ or $139-45$ or $139-46$ or $139-47$ or $139-48$ or $139-49$ or $139-50$ or $139-51$ or $139-52$ or $139-53$ or \ldots \ (incomplete reasoning) \\
\midrule
\textbf{Model:} Qwen2-72B \\
\midrule
\textbf{16k Solution:} \\
We can think of this problem as finding the remainder when $139$ is divided by $11$. Dividing $139$ by $11$, we get a quotient of $12$ and a remainder of $7$. This means that after taking $139$ steps, Tim will have completed $12$ full circles of $11$ steps each, and will be on the $7^\text{th}$ step of the next circle. Therefore, Tim ends on step $\boxed{7}$. \\
\midrule
\textbf{Solution at 128k:} \\
We can write $139 \equiv 3 \pmod{11}$, so Tim ends on step $\boxed{3}$. \\
\midrule
\textbf{Analysis:} \\
Both models correctly solved the problem in their 16k solutions. However, the 128k solution from Llama-3.1-70B devolves into a repetitive and incorrect calculation process, while Qwen2-72B's 128k solution misapplies modular arithmetic, leading to an incorrect result. \\
\bottomrule
\caption{An example problem from MATH-number\_theory}
\label{tbl:math_number_theory}
\end{longtable}
\twocolumn

\onecolumn
\begin{longtable}{p{15cm}}
\small \\
\toprule
\textbf{Task:} BBH-dyck\_languages \\
\midrule
\textbf{Problem:} \\
Complete the rest of the sequence, making sure that the parentheses are closed properly.\\[0.5em]
Input: \verb |<<{{{<[(<>{()})]>}}}| \\
\midrule
\textbf{Difficulty:} Level 4 \\
\midrule
\textbf{Solution:} \\
\verb|> >| \\
\midrule
\textbf{Model:} Llama-3.1-70B \\
\midrule
\textbf{16k Solution:} \\
\verb|> >| \\
\midrule
\textbf{Solution at 128k:} \\
\verb|> >| \\
\midrule
\textbf{Model:} Qwen2-72B \\
\midrule
\textbf{16k Solution:} \\
\verb|> >| \\
\midrule
\textbf{Solution at 128k:} \\
\verb|> >| \\
\midrule
\textbf{Analysis:} \\
Both models, Llama-3.1-70B and Qwen2-72B, consistently provide the same answer \verb|> >| for all context lengths. Their responses correctly complete the given Dyck language sequence by closing the parentheses appropriately.
\\
\bottomrule
\caption{An example problem from BBH-dyck\_languages}
\label{tbl:bbh_dyck_1}
\end{longtable}
\twocolumn
\onecolumn
\begin{longtable}{p{15cm}}
\small \\
\toprule
\textbf{Task:} BBH-dyck\_languages \\
\midrule
\textbf{Problem:} \\
Complete the rest of the sequence, making sure that the parentheses are closed properly.\\[0.5em]
Input: \verb |[<<[{<>([{}{{<>}{}}])}[[[([()[[{<[{{}}]>]>}]]])<<[[(<<()>>)]>>[]]]<([])>| \\
\verb|{((<{}>))}>>| \\
\midrule
\textbf{Difficulty:} Level 5 \\
\midrule
\textbf{Solution:} \\
\verb|]| \\
\midrule
\textbf{Model:} Llama-3.1-70B \\
\midrule
\textbf{16k Solution:} \\
\verb|]| \\
\midrule
\textbf{Solution at 128k:} \\
\verb|> )| \\
\midrule
\textbf{Model:} Qwen2-72B \\
\midrule
\textbf{16k Solution:} \\
\verb|]| \\
\midrule
\textbf{Solution at 128k:} \\
\verb|]| \\
\midrule
\textbf{Analysis:} \\
The expected correct solution to complete the sequence is \verb|]|. Both models provide the correct answer in their 16k solutions. However, at 128k, Llama-3.1-70B deviates from its earlier correct response and outputs \verb|> )|, which is incorrect. Qwen2-72B remains consistent across both context lengths by providing the correct solution \verb|]|. \\
\bottomrule
\caption{An example problem from BBH-dyck\_languages}
\label{tbl:bbh_dyck_2}
\end{longtable}
\twocolumn

\onecolumn
\begin{longtable}{p{15cm}}
\small \\
\toprule
\textbf{Task:} BBH-dyck\_languages \\
\midrule
\textbf{Problem:} \\
Complete the rest of the sequence, making sure that the parentheses are closed properly.\\[0.5em]
Input: \verb|[<<[[](){}{<>([{}{{<>}{}}])}[[[([()[[{<[{{}}<{{<()>}}>]>}]]])<<[[(<<()>>)]| \\
\verb|>>[]]]<([])>{((<{}>))}>>| \\
\midrule
\textbf{Difficulty:} Level 5 \\
\midrule
\textbf{Solution:} \\
\verb|]| \\
\midrule
\textbf{Model:} Llama-3.1-70B \\
\midrule
\textbf{16k Solution:} \\
\verb|]| \\
\midrule
\textbf{Solution at 128k:} \\
\verb|> )| \\
\midrule
\textbf{Model:} Qwen2-72B \\
\midrule
\textbf{16k Solution:} \\
\verb|]| \\
\midrule
\textbf{Solution at 128k:} \\
\verb|]| \\
\midrule
\textbf{Analysis:} \\
The expected correct solution to complete the sequence is \verb|]|. Both models provide the correct answer in their 16k solutions. However, at 128k, Llama-3.1-70B deviates from its earlier correct response and outputs \verb|> )|, which is incorrect. Qwen2-72B remains consistent across both context lengths by providing the correct solution \verb|]|. \\
\bottomrule
\caption{An example problem from BBH-dyck\_languages}
\label{tbl:bbh_dyck_3}
\end{longtable}
\twocolumn

\onecolumn
\begin{longtable}{p{15cm}}
\small \\
\toprule
\textbf{Task:} BBH-word\_sorting \\
\midrule
\textbf{Problem:} \\
Sort the following words alphabetically: \\
List: sediment varsity towel fracture \\
\midrule
\textbf{Difficulty:} Level 1 \\
\midrule
\textbf{Solution:} \\
frature sediment towel varsity \\
\midrule
\textbf{Model:} Llama-3.1-70B \\
\midrule
\textbf{16k Solution:} \\
fracture sediment towel varsity \\
\midrule
\textbf{Solution at 128k:} \\
fracture sediment towel varsity \\
\midrule
\textbf{Model:} Qwen2-72B \\
\midrule
\textbf{16k Solution:} \\
fracture sediment towel varsity \\
\midrule
\textbf{Solution at 128k:} \\
fracture sediment towel varsity \\
\midrule
\textbf{Analysis:} \\
Both models, Llama-3.1-70B and Qwen2-72B, consistently produce the correct alphabetical order: \texttt{fracture sediment towel varsity}, regardless of the context length (16k or 128k). Their responses are accurate and demonstrate stable performance for this word-sorting task.
\\
\bottomrule
\caption{An example problem from BBH-word\_sorting}
\label{tbl:bbh_word_1}
\end{longtable}
\twocolumn

\onecolumn
\begin{longtable}{p{15cm}}
\small \\
\toprule
\textbf{Task:} BBH-word\_sorting \\
\midrule
\textbf{Problem:} \\
Sort the following words alphabetically: \\
List: papaw perfidious frugal labial feel sonar blunderbuss box dinnertime oresteia \\
\midrule
\textbf{Difficulty:} Level 3 \\
\midrule
\textbf{Solution:} \\
blunderbuss box dinnertime feel frugal labial oresteia papaw perfidious sonar \\
\midrule
\textbf{Model:} Llama-3.1-70B \\
\midrule
\textbf{16k Solution:} \\
blunderbuss box dinnertime feel frugal labial oresteia papaw perfidious sonar \\
\midrule
\textbf{Solution at 128k:} \\
box dinnertime labial oresteia papaw perfidious frugal labial sonar perfidious \\
\midrule
\textbf{Model:} Qwen2-72B \\
\midrule
\textbf{16k Solution:} \\
blunderbuss box dinnertime feel frugal labial oresteia papaw perfidious sonar \\
\midrule
\textbf{Solution at 128k:} \\
blunderbuss box dinnertime feel frugal labial oresteia papaw \\
\midrule
\textbf{Analysis:} \\
Both models correctly produce the alphabetical ordering in their 16k solutions: \texttt{blunderbuss box dinnertime feel frugal labial oresteia papaw perfidious sonar}. \\
However, at the 128k context length, Llama-3.1-70B returns an incorrect and inconsistent ordering (\texttt{box dinnertime labial oresteia papaw perfidious frugal labial sonar perfidious}), while Qwen2-72B provides an incomplete solution (\texttt{blunderbuss box dinnertime feel frugal labial oresteia papaw}). \\
\bottomrule
\caption{An example problem from BBH-word\_sorting}
\label{tbl:bbh_word_2}
\end{longtable}
\twocolumn

\onecolumn
\begin{longtable}{p{15cm}}
\small \\
\toprule
\textbf{Task:} BBH-word\_sorting \\
\midrule
\textbf{Problem:} \\
Sort the following words alphabetically: \\
List: logarithm robertson inanimate uganda micron incurring wilcox lumber sammy souvenir newman profuse maladapt cone butch airlift homeowner \\
\midrule
\textbf{Difficulty:} Level 4 \\
\midrule
\textbf{Solution:} \\
airlift butch cone homeowner inanimate incurring logarithm lumber maladapt micron newman profuse robertson sammy souvenir uganda wilcox \\
\midrule
\textbf{Model:} Llama-3.1-70B \\
\midrule
\textbf{16k Solution:} \\
airlift butch cone homeowner inanimate incurring logarithm lumber maladapt micron newman profuse robertson sammy souvenir uganda wilcox \\
\midrule
\textbf{Solution at 128k:} \\
airlift cone butch inanimate homeowner incurring logarithm micron newman profuse robertson uganda wilcox \\
\midrule
\textbf{Model:} Qwen2-72B \\
\midrule
\textbf{16k Solution:} \\
airlift butch cone homeowner inanimate incurring logarithm lumber maladapt micron newman profuse robertson sammy souvenir uganda wilcox \\
\midrule
\textbf{Solution at 128k:} \\
airlift butch cone homeowner inanimate incurring logarithm maladapt micron newman profuse robertson sammy souvenir uganda wilcox \\
\midrule
\textbf{Analysis:} \\
Both models correctly output the full alphabetical ordering in their 16k solutions: \texttt{airlift butch cone homeowner inanimate incurring logarithm lumber maladapt micron newman profuse robertson sammy souvenir uganda wilcox}. \\
\\
At 128k, Llama-3.1-70B produces an output that is missing several words (e.g., \texttt{lumber}, \texttt{sammy}, and \texttt{souvenir}) and reorders some items incorrectly, while Qwen2-72B's 128k solution is mostly correct but also omits \texttt{lumber} compared to the expected solution. \\
\\
These discrepancies indicate that both models encounter difficulties maintaining full consistency at the extended 128k context length for this word sorting task.
\\
\bottomrule
\caption{An example problem from BBH-word\_sorting}
\label{tbl:bbh_word_3}
\end{longtable}
\twocolumn

\onecolumn
\begin{longtable}{p{15cm}}
\small \\
\toprule
\textbf{News Article:} \\
'Philip Bevington, of Camborne, admitted two counts of cultivating cannabis at Truro Crown Court. He told police they were for medicinal purposes to alleviate pain from kidney failure, after being caught by officers twice in two months. Bevington was sentenced to a two-year conditional discharge by Judge Simon Carr, who ordered the drugs destroyed. Officers went to Bevington's home to investigate a "strong smell of cannabis" in November 2019, the court heard. The 80-year-old showed them a sophisticated hydroponic growing set up in his garage housing 30 plants, including a ventilation system, lamps hanging from his ceiling and plastic sheets lining the walls. These plants had the potential to grow three crops a year, which would have had a street selling value of £12,600 to £37,800, the court was told. The same police officers returned two months later in January 2020 and found 44 plants. Judge Simon Carr acknowledged Bevington was growing the cannabis for his "own personal use to alleviate pain", while suffering from his condition and mobility issues, which left him housebound. Related Internet Links HM Courts \& Tribunals Service - GOV.UK \\
\midrule
\textbf{Reference Summary:} \\
'An 80-year-old man caught twice by police with dozens of cannabis plants claimed he used them to treat pain.' \\
\midrule
\textbf{Model:} Llama-3.1-70B \\
\midrule
\textbf{16k Summary:} \\
An 80-year-old man has been given a two-year conditional discharge for growing cannabis for medicinal purposes. \\
\midrule
\textbf{Summary at 128k:} \\
A man in Camborne has been sentenced to a two-year conditional discharge for cultivating cannabis. He told police the plants were for medicinal purposes to alleviate pain from kidney failure. \\
\midrule
\textbf{Model:} Qwen2-72B \\
\midrule
\textbf{16k Summary:} \\
An 80-year-old man who grew cannabis to alleviate pain from kidney failure has been given a conditional discharge. \\
\midrule
\textbf{Summary at 128k:} \\
An 80-year-old man who grew cannabis to ease the pain of kidney failure has been given a conditional discharge. \\
\midrule
\textbf{Analysis:} \\
Llama-3.1-70B's 128k output provides additional location context ("Camborne") and repeats the reason for cultivation, while Qwen2-72B maintains consistency between the 16k and 128k outputs with only minor stylistic differences in phrasing. \\
\bottomrule
\caption{An example news article generation from Llama-3.1-70B and Qwen2-72B}
\label{tbl:xlsum_example}
\end{longtable}
\twocolumn

\end{document}